\documentclass[10pt]{article}
\makeatletter
\providecommand{\@LN}[2]{}
\providecommand{\@LN@col}[1]{}
\makeatother
\ifdefined\pdfminorversion\pdfminorversion=7\fi
\usepackage{fullpage,parskip}
\usepackage[T1]{fontenc}
\usepackage{lmodern}
\usepackage{amsmath,amssymb}
\usepackage{graphicx,xcolor}
\usepackage{booktabs,array,multirow}
\usepackage{subcaption,wrapfig,float}
\usepackage{algorithm,algpseudocode}
\usepackage{needspace}
\usepackage[section]{placeins}
\usepackage[hidelinks]{hyperref}

\newcommand{\papertitle}{Shot-Aware Frame Sampling for Video Understanding}
\title{\papertitle}
\date{}
\author{Mengyu Zhao$^{1,2}$\footnotemark[1] \footnotemark[2] $\quad$ Di Fu$^{1}$\footnotemark[1] $\quad$ Yongyu Xie$^{3}$ $\quad$ Jiaxing Zhang$^{1}$ \\ Zhigang Yuan$^{1}$ $\quad$ Shirin Jalali$^{2}$ $\quad$ Yong Cao$^{1}$
\\
~
\\$^1$ ByteDance $\quad^2$Rutgers University $\quad^3$Georgia Tech
}

\begin{document}
\maketitle

\renewcommand{\thefootnote}{\fnsymbol{footnote}}
\footnotetext[1]{indicates equal contributions. This work was done when Mengyu Zhao was an intern at ByteDance.}
\footnotetext[2]{indicates corresponding author. E-mail: \href{mailto:mykzhao@gmail.com}{mykzhao@gmail.com}\\
Code is available at \url{https://github.com/mengyu02/InfoShot}.}

\begin{abstract}
Video frame sampling is essential for efficient long-video understanding with Vision-Language Models (VLMs), since dense inputs are costly and often exceed context limits.
Yet when only a small number of frames can be retained, existing samplers often fail to balance broad video coverage with brief but critical events, which can lead to unreliable downstream predictions.
To address this issue, we present {InfoShot}, a task-agnostic, shot-aware frame sampler for long-video understanding.
{InfoShot} first partitions a video into semantically consistent shots, and then selects two complementary keyframes from each shot: one to represent the main content and one to capture unusual within-shot changes.
This design is guided by an information-theoretic objective that encourages the sampled set to retain high information about both dominant shot content and sparse within-shot events.
In this way, it improves the chance of preserving both overall video context and short decision-critical moments without requiring any retraining.
To better evaluate such short-lived events, we further introduce SynFlash, a synthetic benchmark with controllable sub-second anomaly patterns and frame-level ground truth, and we also evaluate InfoShot on existing anomaly datasets and general video understanding tasks.
Experiments show that InfoShot improves anomaly hit rate and downstream Video-QA accuracy under frame number constraints, while matching or outperforming strong baselines on standard video understanding benchmarks.

\end{abstract}

\section{Introduction}
\label{sec:intro}
\begin{figure}[htbp]
    \centering
    \includegraphics[width=0.99\linewidth]{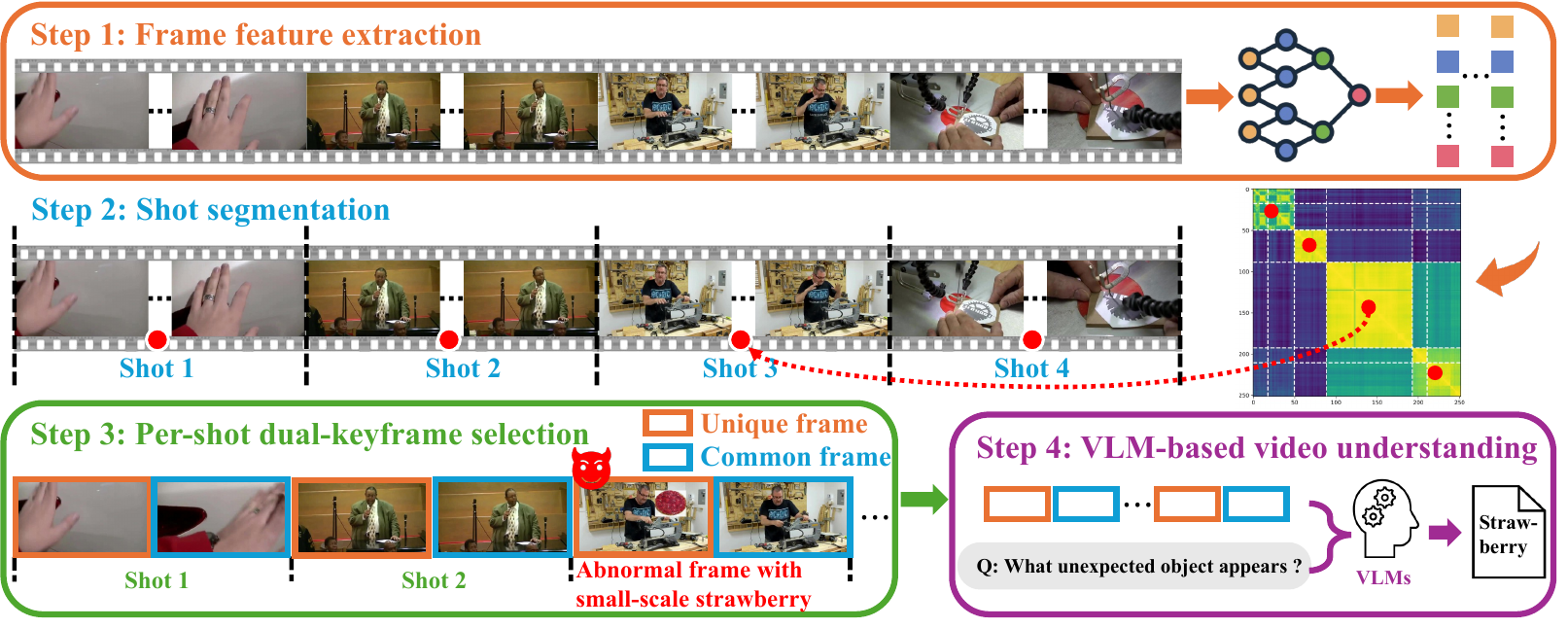}
    \caption{InfoShot overview. Under a fixed budget, greedy shot segmentation (Sec.~\ref{subsec:shot_seg}) precedes selection of a representative \emph{common} frame and a high-deviation \emph{unique} frame per shot (Sec.~\ref{subsec:dualframe}). The strawberry example uses picture-in-picture (PiP) injection into a small local region. All samplers use the same VLM prompt.}
    \label{fig:framework}
    \end{figure}

Vision-Language Models (VLMs) have substantially improved video understanding and reasoning~\cite{gpt4v,gemini,qwenvl,Qwen2.5-VL,lin2024video,videochatgpt,chen2024internvl}. However, processing dense spatiotemporal inputs remains expensive and often exceeds the context limits of current architectures~\cite{wu2024longvideobench,jin2024chat,li2024mvbench,fu2025video,yu2024frame,li2025improving}. To make long-video inference practical, existing systems mainly rely on in-model token compression~\cite{bolya2022token,chen2024image,ryoo2021tokenlearner,rao2021dynamicvit,Norouzi_2024_ALGM} or front-end frame sampling~\cite{wu2019adaframe,yu2023self,korbar2019scsampler,ghodrati2021frameexit,yu2024frame,yao2025generative}. This paper focuses on frame sampling. Sampling fixes the visual evidence available to the downstream VLM; once a critical frame is missed, the model cannot recover it, which can lead to unsupported or wrong Video-QA and reasoning outputs~\cite{wu2024longvideobench,li2024mvbench,fu2025video,yu2024frame}.

A key challenge is that information in real videos is not evenly distributed over time. Uniform temporal sampling~\cite{wang2018temporal} allocates frames by timestamps and can oversample long static shots with near-duplicate frames while undersampling fast motions and short events~\cite{wu2019adaframe,korbar2019scsampler,yu2024frame,yao2025generative}. Other approaches attempt to rank frames using task-specific scores (e.g., anomaly scores) or query-conditioned matching for Video-QA/retrieval~\cite{wu2024vadclip,zhang2025holmes,tang2025adaptive,yu2024frame,zhang2025q,wu2023empirical}. These methods often require supervision, rely on a known query at sampling time, or depend on scoring signals that may not generalize. Under tight frame budgets, they can still miss sub-second but decision-critical evidence~\cite{li2025improving,zhao2025efficient}. This motivates \emph{shot-aware, content-driven} sampling: instead of sampling by time, we segment videos into semantically consistent shots and allocate the budget across shots, so that sampling density follows visual content~\cite{soucek2024transnet,zhu2023autoshot}.

This issue is especially important for tasks that require evidence of brief events, such as anomaly detection~\cite{sultani2018real,wu2020not} and Video-QA about transient content. Many anomalies last only a fraction of a second in real videos. If the anomaly frame is not sampled, even strong VLMs cannot infer it reliably from surrounding frames. In practice, transient anomalies are also diverse: beyond black/white frames, they include high-saturation flashes, short wipe-like transitions, picture-in-picture artifacts with small spatial support, and camouflaged insertions with matched global color statistics. These patterns challenge global similarity measures and threshold-based heuristics~\cite{pyscenedetect,hassanien2017large,soucek2024transnet}, which can treat localized or low-contrast changes as minor variation.

To address this bottleneck, we propose \textbf{InfoShot}, a task-agnostic, model-independent sampling framework guided by an information-theoretic view: under a fixed budget, the sampled set should retain high information about latent video content, including dominant shot semantics and sparse within-shot events. We model a video as a locally stationary background process within each segment, with sparse deviations for transient events. Motivated by this model, InfoShot performs greedy shot segmentation using a windowed feature contrast, and then selects two complementary keyframes per shot: a \emph{common} frame that represents the dominant semantics and a \emph{unique} frame selected using atypicality and local feature changes. This design preserves global context, increases the chance of capturing sub-second evidence, and reduces redundancy in long static shots.

Evaluating short-lived events is difficult because existing benchmarks often lack frame-level ground truth for transient anomalies. We therefore introduce \textbf{SynFlash}, a synthetic Video-QA benchmark designed to reflect real-world moderation settings with controllable sub-second anomaly patterns and exact frame-level annotations. We further evaluate on real-world anomaly datasets (HIVAU-70k~\cite{zhang2025holmes}) and a general video reasoning benchmark (Video-MME~\cite{fu2025video}). Experiments show that InfoShot improves anomaly hit rate and downstream Video-QA accuracy under low-frame-rate constraints, while matching or outperforming strong baselines on standard video understanding benchmarks. We deploy InfoShot in TikTok’s video moderation pipeline, which processes billions of short videos per week, and observe substantially higher recall for evasive transient harmful content under tight compute constraints.

Our contributions are as follows:

\begin{itemize}
    \item We propose InfoShot (Fig.~\ref{fig:framework}), a shot-aware sampler that segments videos in deep feature space and selects informative keyframes to improve semantic coverage for general video understanding under a limited frame budget.
    \item We introduce SynFlash, a synthetic Video-QA benchmark designed to reflect real-world moderation settings with four anomaly patterns and frame-level ground truth, and show consistent improvements over task-agnostic sampling baselines under limited frame budgets.
    \item We evaluate on HIVAU-70k~\cite{zhang2025holmes} and Video-MME~\cite{fu2025video}, showing improved anomaly coverage while preserving Video-QA performance under the same sampling rate.
    \end{itemize}

\section{Related Work}
\label{sec:related_work}
\noindent\textbf{Shot Boundary Detection and Scene Segmentation.}
Shot boundary detection (SBD) partitions a video into temporal segments (shots) based on visual transitions. Heuristic methods based on low-level cues are still used in practice (e.g., PySceneDetect~\cite{pyscenedetect}), but often fail in real-world edits. Recent work uses deep models for more reliable transition detection (e.g., TransNet~V2~\cite{soucek2024transnet}) or shot-aware representations (e.g., ShotCoL~\cite{chen2021shot}); AutoShot~\cite{zhu2023autoshot} shows that gradual transitions remain challenging. Temporal action segmentation models~\cite{farha2019ms, yi2021asformer} capture long-range structure but target semantic labeling rather than boundary detection. In this work, we use shot boundaries as macroscopic structure to allocate sampling budget by content rather than by time.

\noindent\textbf{Video Anomaly Detection and Understanding.}
Video anomaly detection (VAD) aims to identify abnormal events, often under weak supervision. Many methods use multiple-instance learning on untrimmed datasets~\cite{sultani2018real, wu2020not} or rely on reconstruction/prediction errors~\cite{gong2019memorizing, liu2018future}. Recent work improves localization by modeling temporal feature magnitude~\cite{tian2021weakly} or aligning features with language~\cite{wu2024vadclip}. Beyond binary detection, benchmarks (e.g., HIVAU-70K~\cite{zhang2025holmes}) emphasize temporal coverage and fine-grained reasoning over long videos. However, many VAD pipelines depend on task-specific anomaly scores for ranking or sampling, which may not generalize to unseen patterns. Under low frame rates, missing a short anomaly can dominate the final error.

\noindent\textbf{Video Frame Sampling.}
Efficient sampling is a common way to reduce compute in video recognition. Uniform sparse sampling~\cite{wang2018temporal} remains a strong baseline, but it allocates frames by time and can oversample long static shots while undersampling fast events. Adaptive methods learn policies to select informative frames or clips under a budget, for example via conditional computation~\cite{wu2019adaframe, korbar2019scsampler} or early exiting~\cite{ghodrati2021frameexit}. In multimodal systems, many samplers are trained or tuned for specific objectives such as retrieval~\cite{hu2022mmsampler}, recognition~\cite{wang2023differentiable}, or Video-QA~\cite{yu2023self,tang2025adaptive}. Among query-based methods, Frame-Voyager~\cite{yu2024frame} learns to select frame combinations for a given question. AKS~\cite{tang2025adaptive} balances question relevance with video coverage, while GenS~\cite{yao2025generative} uses a VideoLLM to identify question-relevant frames. Our setting requires a task-agnostic sampler that preserves both dominant video content and brief but informative events under a limited frame budget, without task-specific supervision or a known query. We therefore use a shot-aware strategy that selects representative and high-deviation frames within each shot to retain complementary visual information.

\noindent\textbf{Video-QA and Vision-Language Models.}
VLMs have been extended to video by encoding a sequence of sampled frames and performing reasoning~\cite{videochatgpt, videochat, zhang2023video}. Recent models scale instruction tuning and improve performance on video benchmarks~\cite{lin2024video, zhang2024llava, qwenvl, li2024mvbench, fu2025video}. But these systems typically treat sampling as a preprocessing step with sufficient evidence. We instead treat frame extraction as a primary failure mode under strict budgets, and we study how to select frames that better cover both context and short-critical events.

\section{Methodology}
\label{sec:methodology}

The overall pipeline is shown in Fig.~\ref{fig:framework}. It is inspired by an information-theoretic view: under a fixed frame budget, the sampled set should retain information about dominant content and sparse events. This suggests preserving both (i) shot-level structure and (ii) sparse within-shot deviations that may carry decision-critical evidence. InfoShot implements this with two stages: it forms a budget-constrained shot partition in deep feature space (Step~2) and then selects two complementary keyframes per shot (Step~3), a \emph{common} frame and a \emph{unique} high-deviation frame. The sampled frames are fed to the downstream VLM with the same prompt for Video-QA (Step~4); only the sampled frame set differs across methods, and the sampler can be used with other VLM-based video tasks. The following sections formalize this principle and describe the InfoShot procedure.

\subsection{Problem Formulation}
Let a video be a sequence of frames $\mathcal{V}=\{x_t\}_{t=1}^{T}$. We extract features using a pretrained visual encoder $\phi$, $c_t=\phi(x_t)\in\mathbb{R}^n$ (Step~1 in Fig.~\ref{fig:framework}). Deep features reduce low-level noise, but can change sharply under transient events.

Given features $\{c_t\}_{t=1}^T$ and a frame budget $K$, we select indices $\mathcal{K}\subset\{1,\dots,T\}$ with $|\mathcal{K}|\le K$. We introduce two latent binary sequences: a shot boundary indicator $s_t\in\{0,1\}$ and a transient event indicator $f_t\in\{0,1\}$. Here $s_t=1$ indicates a shot change at time $t$, and $f_t=1$ indicates a localized within-shot deviation at time $t$. Our goal is to choose $\mathcal{K}$ that preserves both shot-level structure and rare within-shot events under a strict budget.

\subsection{Stochastic Sequence Model}
We use a lightweight generative abstraction for stable background content and sparse visual events. Let $m(t)$ denote the segment containing frame $t$ under the partition induced by $S_{1:T}$. Within segment $m$, the latent background feature is
\begin{equation}
b_t=\mu_m+\epsilon_t,\qquad t\in\mathcal S_m,
\end{equation}
where $\mu_m$ represents dominant content. We approximate the residual process as locally stationary, with Gaussian marginals
\begin{equation}
\epsilon_t\sim\mathcal N(0,\Sigma_m),
\end{equation}
and temporal covariance decaying with lag. The observed feature combines background and event components:
\begin{equation}
c_t=(1-f_t)b_t+f_t\eta_t.
\label{eq:stochastic_model}
\end{equation}
When $f_t=0$, the frame follows the background process; when $f_t=1$, $\eta_t$ represents an event-dependent deviation. Shot boundaries change the background parameters. The stationarity assumption concerns the background component within a segment, while the observed sequence may contain abrupt changes. This abstraction motivates preserving both representative context and sparse deviations; it does not assume that every feature outlier is informative.

\begin{figure}[htbp]
\centering
\includegraphics[width=0.99\linewidth]{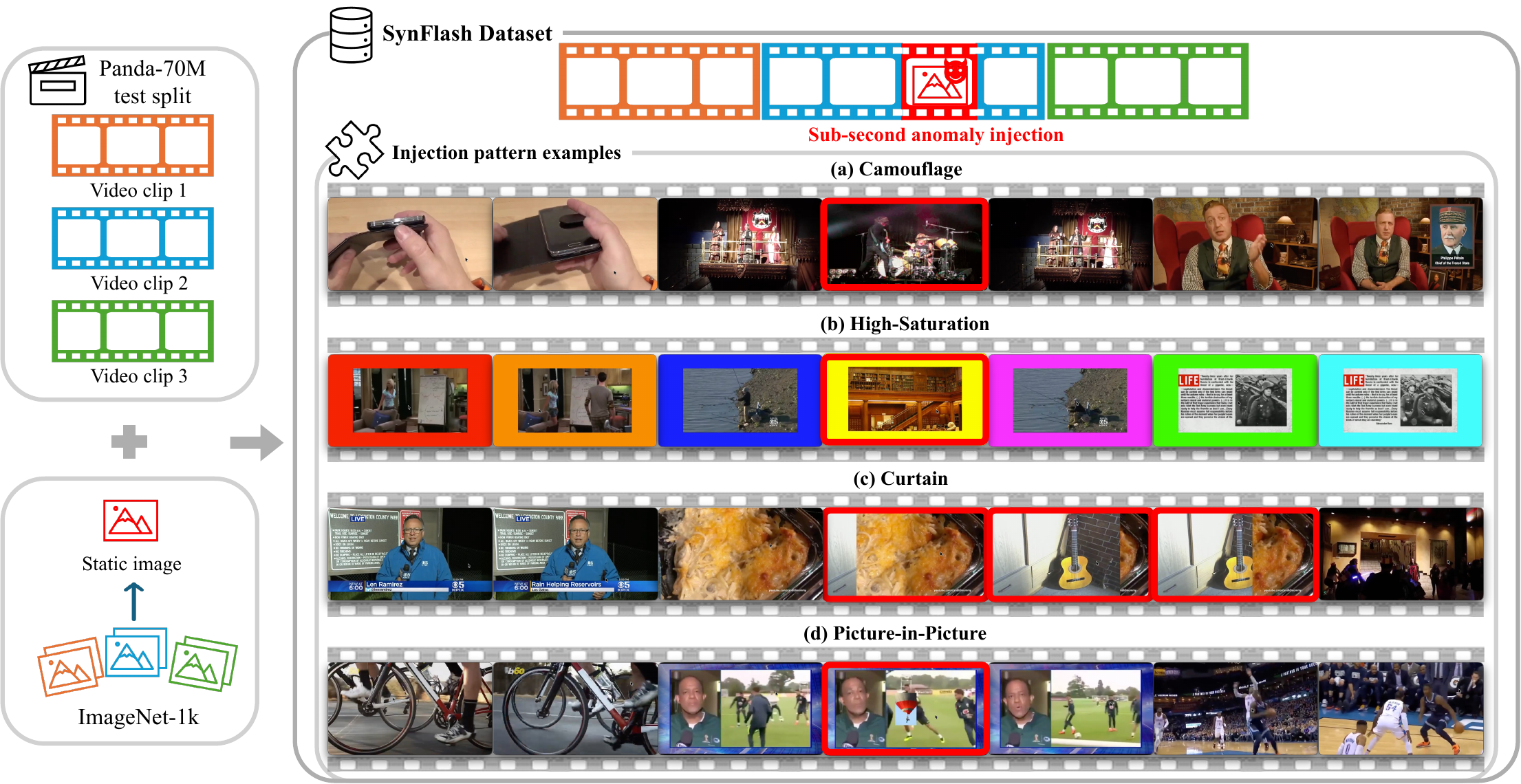}
\caption{Overview of the SynFlash generation pipeline. ImageNet-1k~\cite{imagenet15russakovsky} images are injected at random times into videos formed by concatenating three Panda-70M~\cite{chen2024panda70m} clips, creating four transient anomaly types with frame-level ground truth.}
\label{fig:data_pipeline}
\end{figure}

\subsection{Information-Theoretic Principle}
Our goal is to preserve video information without a known downstream query. Let $G$ represent dominant shot content and temporal structure, including the partition and background parameters, and let $E$ represent sparse-event states, content, and locations. We use the following design objective:
\begin{equation}
\max_{\mathcal K\subset\{1,\dots,T\},\,|\mathcal K|\le K}I(C_{\mathcal K};G,E),
\end{equation}
where $C_{\mathcal K}=\{c_t\}_{t\in\mathcal K}$. The chain rule gives
\begin{equation}
I(C_{\mathcal K};G,E)=I(C_{\mathcal K};G)+I(C_{\mathcal K};E\mid G).
\end{equation}
The first term concerns dominant content and shot structure; the second concerns event evidence beyond that context. This motivates organizing the video into segments and selecting complementary common and unique frames. Repeated representative frames may add little evidence about $E$, whereas a meaningful deviation may supply information missing from the context. Our scores are practical surrogates, not direct mutual-information estimates or a guarantee of optimal selection (Appendix~\ref{sup:info_maximization}).

\subsection{Shot Segmentation via Windowed Contrast}
\label{subsec:shot_seg}
To preserve shot-level context, we construct a budget-constrained partition using windowed feature changes (Step~2 in Fig.~\ref{fig:framework}). Let $A_{ij}=\cos(c_i,c_j)$ be pairwise affinity. For a boundary $t$, define two windows immediately before and after it
$\mathcal{W}_t^-=\{t-k,\dots,t-1\}$ and $\mathcal{W}_t^+=\{t,\dots,t+k-1\}$.
We score boundary strength by comparing within-window cohesion and cross-window similarity:
\begin{equation}
\begin{aligned}
B_t ={}& \tfrac{1}{2}\big(\mathbb{E}_{I_1,I_2}[A_{I_1 I_2}]
               \!+\! \mathbb{E}_{J_1,J_2}[A_{J_1 J_2}]\big)\!-\!\mathbb{E}_{I_1,J_1}[A_{I_1 J_1}] .
\end{aligned}
\label{eq:boundary_score}
\end{equation}
Here $I_1\ne I_2$ and $J_1\ne J_2$ are uniform pairs within $\mathcal W_t^+$ and $\mathcal W_t^-$, respectively; the cross term averages all cross-window pairs. Windows are clipped at segment ends. Appendix~\ref{sup:boundary_score} relates this contrast to mean-change tests.

We greedily split segments based on $B_t$. Starting from one segment, we repeatedly choose the split point with the largest $B_t$ inside current segment until reaching $M=\lfloor K/2\rfloor$ segments, reserving two frames per segment for the next stage. Pseudocode of the full InfoShot pipeline is provided in Appendix~\ref{sup:pseudocode}.

\subsection{Dual-Keyframe Selection Within Each Shot}
\label{subsec:dualframe}
To retain event evidence alongside representative context, we select complementary within-shot frames (Step~3 in Fig.~\ref{fig:framework}). Let shot $m$ contain indices $\mathcal{S}_m$ with length $L=|\mathcal{S}_m|$ and affinity matrix $A\in\mathbb{R}^{L\times L}$ restricted to this shot.

For each frame $i\in\mathcal{S}_m$, we compute two scores: \emph{typicality} as the average similarity to the shot and \emph{local volatility} as the dissimilarity to a temporal neighborhood $\mathcal{N}_i$,
\begin{equation}
\begin{aligned}
g_i = \frac{1}{L-1}\sum_{j\in\mathcal{S}_m,\,j\neq i} A_{ij},\qquad
v_i &= 1-\frac{1}{|\mathcal{N}_i|}\sum_{j\in\mathcal{N}_i} A_{ij}.
\end{aligned}
\end{equation}
We normalize scores within the shot to obtain $\hat g_i,\hat v_i\in[0,1]$.

\noindent\textbf{Common and unique keyframes.}
We select two complementary frames per shot: a \emph{common} frame that is globally typical and locally stable, and a \emph{unique} frame that favors atypicality and local change:
\begin{equation}
\begin{aligned}
t_{\mathrm{com}} &= \operatorname*{arg\,max}_{i\in\mathcal{S}_m}
    \big[\lambda \hat g_i-(1-\lambda)\hat v_i\big],\\
t_{\mathrm{uni}} &= \operatorname*{arg\,max}_{i\in\mathcal{S}_m\setminus\{t_{\mathrm{com}}\}}
    \big[\alpha(1-\hat g_i)+(1-\alpha)\hat v_i\big].
\end{aligned}
\label{eq:uni_com}
\end{equation}
The final keyframe set is the union of $\{t_{\text{com}},t_{\text{uni}}\}$ over all shots, truncated if necessary to satisfy $|\mathcal{K}|\le K$.

\noindent\textbf{Rationale for two keyframes.}
The common frame represents dominant shot semantics, while the unique frame targets sparse within-shot deviations. These intended roles are consistent with the information decomposition above; the detailed per-shot analysis is given in Appendix~\ref{sup:dualframe_rationale}.

\noindent\textbf{Why combine typicality and volatility?}
Typicality alone may select a globally similar but locally unstable transition frame, so the common score penalizes volatility. For unique frames, volatility can favor motion or noise, while atypicality can miss localized, smooth evidence. Combining them balances these cues. Section~\ref{subsec:frame_roles} tests the unique frame; Appendix~\ref{sup:parameters} reports score-weight sensitivity.

\noindent\textbf{Sampling cost.}
InfoShot uses an encoder and no VLM calls during selection. Appendix~\ref{sup:complexity} separates feature extraction, affinity computation, and split search; arithmetic bounds alone do not establish a runtime advantage.

\section{Experiments}
\label{sec:experiments}
We evaluate the proposed frame sampling method on both synthetic and real-world benchmarks. We first introduce SynFlash, a synthetic Video-QA dataset tailored for sub-second anomalies with exact frame-level ground truth. We then evaluate event-level coverage on HIVAU-70k ~\cite{zhang2025holmes} annotations and test general video understanding on Video-MME~\cite{fu2025video}. We compare methods under a fixed sampling rate (fps) budget rather than a fixed number of frames in all tests.

\subsection{Dataset Construction: SynFlash}
\label{subsec:synflash}
To evaluate whether a sampler preserves sub-second visual evidence in moderation settings, we construct \textbf{SynFlash}, a synthetic \emph{flash-frame} dataset, via controlled anomaly injection. The injection patterns are based on flash-frames we observed in real-world moderation data, and are designed to reflect harmful cases where short but important evidence is likely to be dropped. As illustrated in Fig.~\ref{fig:data_pipeline}, each synthetic video is composed of three background clips sampled from the Panda-70M~\cite{chen2024panda70m} test split and concatenated to keep the total duration under 30 seconds. We normalize all videos to a fixed resolution of $640 \times 480$ and a fixed frame rate of 24 fps. We then inject short anomalies using images randomly drawn from ImageNet-1k~\cite{imagenet15russakovsky} following four injection patterns, and record the injection timestamps and ImageNet labels for evaluation.

\noindent\textbf{Injection protocol.}
For each video, we sample an insertion time uniformly within the valid temporal range and inject a sub-second anomaly segment whose duration depends on the injection pattern. We generate four subsets to cover common patterns:
(a) \textbf{Camouflage}: we apply a color statistics matching transform in CIELAB space to make the inserted image match the local background appearance before blending, reducing low-level contrast;
(b) \textbf{High-saturation}: we simulate flash-like distractors by applying high fps color perturbations on the background (e.g., a flashing ring) for the whole video;
(c) \textbf{Curtain}: we simulate rapid wipe-like edits by sliding in the anomaly around the injection time, optionally with brief blackout frames to break temporal continuity;
(d) \textbf{Picture-in-picture (PiP)}: we downscale the inserted image and paste it into a random local region to test anomalies with small spatial support.
In total, SynFlash contains 8{,}000 videos (2{,}000 per subset).

\subsection{Benchmark Design and Video-QA Protocol}
\label{subsec:benchmark_design}

\noindent\textbf{Evaluation setup.}
To evaluate performance on capturing short-critical events, we use SynFlash and compare model-independent samplers under the same time-based budget of 0.5 fps. For a video of duration $L$ seconds, the target number of sampled frames is $K=\lceil 0.5L\rceil$, and each method outputs a set of frame indices $\mathcal{K}$ with $|\mathcal{K}|\le K$. We compare against three baselines.
\textbf{Uniform} samples frames at fixed temporal intervals.
\textbf{VSUMM}~\cite{de2011vsumm} is an unsupervised summarization baseline based on clustering; we set the target number of selected keyframes to $K$ and map each selected representative to its nearest original frame.
\textbf{TransNet} (TransNet~V2~\cite{soucek2024transnet}) predicts frame-level transition scores; we rank frames by the score and select the top-$K$ frames.

\noindent\textbf{Implementation details.}
We compute frame features using a pretrained ResNet-50 with the classification head removed; we use the standard preprocessing and apply $\ell_2$ normalization, and all similarity matrices use cosine similarity between normalized features. For InfoShot, greedy shot segmentation uses a boundary window size in the range 3-15 (scaled by segment length) with shot-length constraints of 3-300 frames; within each shot we select two keyframes using temporal neighborhood size $k=1$, $\lambda=0.7$ (common) and $\alpha=0.5$ (unique). For efficiency on long videos, we compute similarity in blocks of 600 frames at 10 fps. All sampled frames are fed to {Qwen2.5-VL}~\cite{Qwen2.5-VL} using the same prompt template and deterministic decoding (temperature 0); the only difference across methods is the sampled frame set.

\begin{figure}[htbp]
    \centering
    \includegraphics[width=1\linewidth]{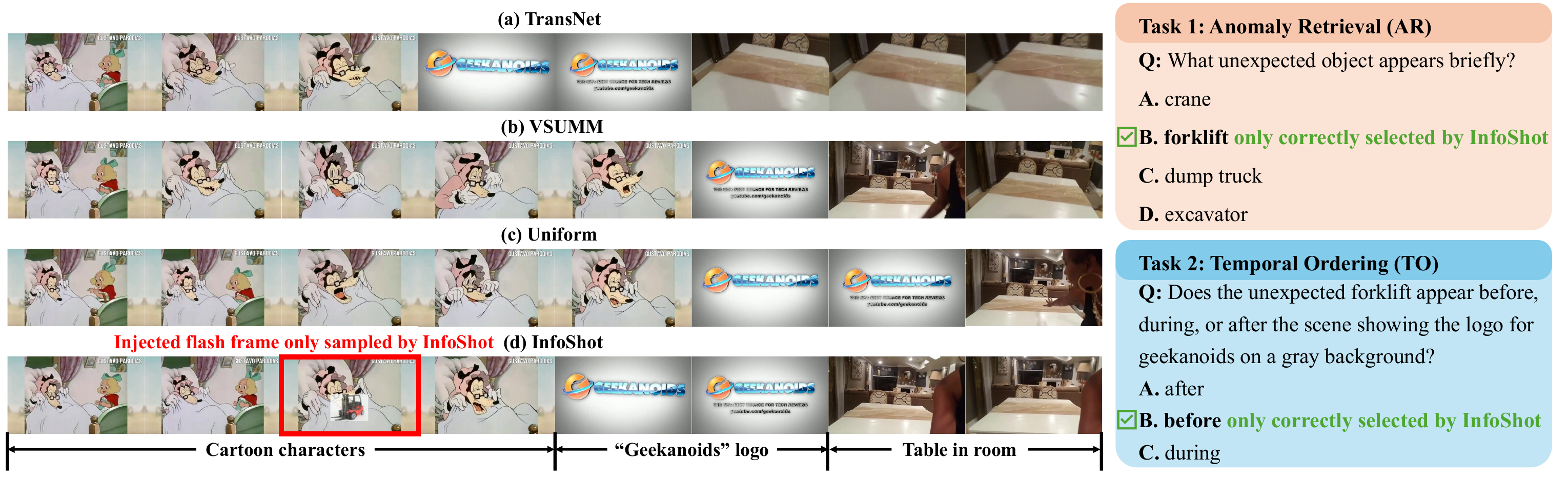}
    \caption{SynFlash picture-in-picture (PiP) case, with a forklift inserted into a small local region. Left: sampled frames from different methods. Right: example questions.}
    \label{fig:vqa_showcase}
\end{figure}

\begin{table}[htbp]
    \centering
    \caption{Performance on SynFlash. We report Anomaly Hit Rate (R, \%), Anomaly Retrieval Accuracy (AR, \%), and Temporal Ordering Accuracy (TO, \%). Best results are \textbf{bold}; second-best are \underline{underlined}.}
    \label{tab:main_results}
    \small
    \setlength{\tabcolsep}{2pt}
    \begin{tabular*}{\linewidth}{@{\extracolsep{\fill}}lllllll@{}}
        \toprule
        \textbf{Metric} & \textbf{Method} & \textbf{Camou.} & \textbf{Hi-Sat.} & \textbf{Curt.} & \textbf{PiP} & \textbf{Avg.} \\
        \midrule
        \multirow{4}{*}{\textbf{R}}
        & Uniform  & 11.20 & 10.00 & 19.15 & \underline{11.20} & 12.89 \\
        & VSUMM    & 74.50 & 9.15 & \textbf{93.75} & 9.35 & 46.69 \\
        & TransNet & \underline{95.55} & \underline{16.55} & 81.15 & 10.30 & \underline{50.89} \\
        & \textbf{InfoShot} & \textbf{98.35} & \textbf{87.05} & \underline{88.45} & \textbf{64.95} & \textbf{84.70} \\
        \midrule
        \multirow{4}{*}{\textbf{AR}}
        & Uniform  & 29.60 & 29.05 & 34.70 & 29.75 & 30.78 \\
        & VSUMM    & 56.75 & 28.40 & \textbf{74.95} & 29.45 & \underline{47.39} \\
        & TransNet & \underline{62.70} & \underline{30.80} & 60.85 & \underline{30.10} & 46.11 \\
        & \textbf{InfoShot} & \textbf{69.15} & \textbf{70.60} & \underline{73.15} & \textbf{48.35} & \textbf{65.31} \\
        \midrule
        \multirow{4}{*}{\textbf{TO}}
        & Uniform    & 35.00 & 33.85 & 36.80 & 35.35 & 35.25 \\
        & VSUMM      & 53.90 & 33.40 & \underline{62.75} & 34.10 & 46.04 \\
        & TransNet   & \textbf{61.95} & \underline{35.30} & \textbf{66.50} & \underline{35.80} & \underline{49.89} \\
        & \textbf{InfoShot} & \underline{60.70} & \textbf{56.85} & 61.10 & \textbf{52.75} & \textbf{57.85} \\
        \bottomrule
    \end{tabular*}
\end{table}

\noindent\textbf{Metrics.}
We measure whether a sampler captures the injected anomaly and whether the sampled evidence supports downstream reasoning. Let $\mathcal{K}$ denote sampled frame indices and let $\mathcal{A}$ denote the set of injected anomaly frame indices recorded during generation. We report \textbf{Anomaly Hit Rate (R)} as the percentage of videos where $\mathcal{K} \cap \mathcal{A} \neq \emptyset$. We also generate two multiple-choice questions per video using SynFlash metadata, yielding 16{,}000 queries: (i) \textbf{Anomaly Retrieval (AR)} asks for the identity/category of the injected content; for AR, we construct hard distractors by sampling answer choices from the same superclass/type as the injected ImageNet label. (ii) \textbf{Temporal Ordering (TO)} asks the temporal order between the anomaly and a macroscopic event in the background. We report per-task accuracy over all queries.

\noindent\textbf{Results.}
Table~\ref{tab:main_results} reports results by injection pattern subsets. InfoShot achieves the best average anomaly hit rate, AR, and TO accuracy, with per-pattern exceptions. The gains are largest on High-saturation and PiP, where global-similarity and boundary-only strategies often miss short or localized events. Fig.~\ref{fig:vqa_showcase} shows a VQA case where InfoShot samples the flash evidence that other methods miss. When the anomaly is not sampled, the VLM answers without the required evidence. InfoShot selects a representative frame and a high-deviation frame within the shot, improving performance on our metrics. Additional case studies are provided in Appendix~\ref{sup:case_study}.

\begin{figure}[!htbp]
\centering
\includegraphics[width=0.65\linewidth]{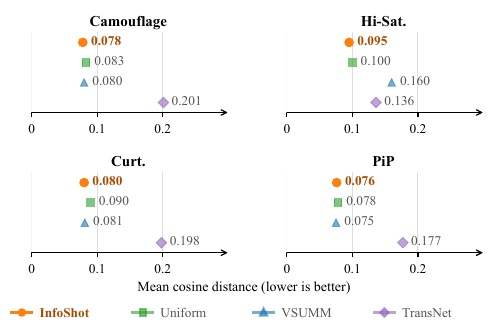}
\caption{SynFlash feature-space coverage at 0.5 fps: mean nearest-frame cosine distance \textbf{(lower is better)}.}
\label{fig:coverage_main}
\end{figure}

\noindent\textbf{Feature-space coverage.}
Figure~\ref{fig:coverage_main} compares coverage of the full video at 0.5 fps. With cosine affinity $A_{tk}$, we measure the mean nearest-frame distance in the shared feature space:
\begin{equation}
\mathrm{Dist}(\mathcal{K})=\frac{1}{T}\sum_{t=1}^{T}\min_{k\in\mathcal{K}}(1-A_{tk}).
\label{eq:set_distortion}
\end{equation}
InfoShot has the lowest distortion on Camouflage, High-saturation, and Curtain. On PiP, VSUMM is slightly lower (0.075 vs.\ 0.076), despite InfoShot's higher anomaly hit rate. Coverage and event capture therefore measure different properties: long background segments can dominate the average distance. This feature-space diagnostic neither estimates mutual information nor establishes the common frame's causal contribution. Appendix~\ref{sup:distortion} provides details.

\subsection{Real-World Anomaly Detection}
\label{subsec:real_world_anomaly}

SynFlash provides exact frame-level ground truth for transient anomalies. We next test whether InfoShot improves event coverage on real videos using HIVAU-70K~\cite{zhang2025holmes} annotations over UCF-Crime~\cite{sultani2018real} and XD-Violence~\cite{wu2020not}. We impose an extreme sampling rate of 0.1 fps for all methods. We evaluate event-level hit metrics computed directly from HIVAU-70K annotation timestamps.

\noindent\textbf{Metrics.}
Each video contains annotated event intervals $\{[a_i,b_i]\}$ in seconds. We map timestamps to frame indices using the decoded frame rate and report:
(i) \textbf{Event Recall (ER)}: the fraction of events for which at least one sampled frame falls within $[a_i,b_i]$;
(ii) \textbf{Complete Coverage Rate (CR)}: the fraction of videos for which each annotated event is hit by at least one sampled frame.

\noindent\textbf{Results.}
Table~\ref{tab:general_anomaly} shows that InfoShot achieves the best overall coverage among the compared methods under an extremely tight frame budget. It improves both ER and CR on XD-Violence and improves CR on UCF-Crime while matching the best ER. Uniform remains competitive on UCF-Crime ER, which is consistent with longer event durations where periodic sampling can still land inside event intervals.

\begin{table}[htbp]
    \centering
    \caption{Event coverage on HIVAU-70k annotations under 0.1 fps. ER: event recall, CR: complete coverage rate. Weighted Avg. is weighted by the number of events/videos in each dataset. Best results are highlighted in \textbf{bold}, and second-best are \underline{underlined}.}
    \label{tab:general_anomaly}
    \small
    \setlength{\tabcolsep}{3.5pt}
    \begin{tabular}{@{} l ll ll ll @{}}
        \toprule
        \multirow{2}{*}{\textbf{Method}} & \multicolumn{2}{l}{\textbf{UCF-Crime}} & \multicolumn{2}{l}{\textbf{XD-Violence}} & \multicolumn{2}{l}{\textbf{Wtd. Avg.}} \\
        \cmidrule(lr){2-3} \cmidrule(lr){4-5} \cmidrule(l){6-7}
        & \textbf{ER} & \textbf{CR} & \textbf{ER} & \textbf{CR} & \textbf{ER} & \textbf{CR} \\
        \midrule
        TransNet            & 40.97 & 37.50 & 37.16 & 39.00 & 37.56 & 38.69 \\
        VSUMM               & \underline{65.97} & 62.50 & 57.35 & 51.60 & 58.25 & 53.82 \\
        Uniform             & \textbf{69.44} & \underline{65.62} & \underline{59.53} & \underline{54.20} & \underline{60.56} & \underline{56.53} \\
        \textbf{InfoShot}       & \textbf{69.44} & \textbf{66.41} & \textbf{63.09} & \textbf{58.40} & \textbf{63.75} & \textbf{60.03} \\
        \bottomrule
    \end{tabular}
    \end{table}

\subsection{General Video Understanding}
\label{subsec:videomme}

\begingroup
\setlength{\columnsep}{8pt}
\setlength{\intextsep}{4pt}
\begin{wraptable}{r}{0.50\columnwidth}
    \centering
    \caption{Video-MME accuracy (\%) at 0.25 fps. Best: \textbf{bold}; second-best: \underline{underlined}.}
    \label{tab:videomme_mini}
    \small
    \setlength{\tabcolsep}{3pt}
    \begin{tabular}{@{} l l l l @{}}
        \toprule
        \textbf{Method} & \textbf{Short} & \textbf{Med.} & \textbf{Avg.} \\
        \midrule
        TransNet             & 65.8 & 60.4 & 63.2 \\
        VSUMM                & 67.8 & \textbf{62.5} & \underline{65.2} \\
        Uniform              & \textbf{71.8} & 62.2 & \textbf{67.0} \\
        \textbf{InfoShot} & \underline{71.5} & \underline{62.4} & \textbf{67.0} \\
        \bottomrule
    \end{tabular}
    \end{wraptable}
Finally, we test whether a sampler designed to preserve transient anomalies harms general video reasoning. We evaluate on Video-MME~\cite{fu2025video} with the same fixed sampling rate for all methods (0.25 fps). We report accuracy on short and medium videos following the benchmark protocol.

\noindent\textbf{Results.} As shown in Table~\ref{tab:videomme_mini}, InfoShot matches uniform sampling on average and is slightly better on medium videos under the same frame sampling budget. This suggests that the shot-aware allocation does not reduce the evidence needed by Video-MME, while providing better coverage for transient anomalies in SynFlash and HIVAU-70K.

\par\endgroup

\subsection{Comparison with Query-Based Samplers}
\label{subsec:query_based}
\noindent\textbf{Evaluation setup.} We compare recent query-based samplers AKS-SeViLA~\cite{tang2025adaptive,yu2023self} and GenS~\cite{yao2025generative} using their released frame-selection models: the SeViLA scorer for AKS and Qwen2.5-VL-3B for GenS. We evaluate on SynFlash, HIVAU, and the short and medium splits of Video-MME, using the same sampling-rate budgets across methods: 0.5, 0.1, and 0.25 fps, respectively. To match our setting, AKS (fixed) and GenS (fixed) use the same generic prompt without access to the downstream question or answer choices. Each method retains its own prompt template and frame-selection rules. On SynFlash, their AR/TO variants receive the corresponding task question and answer choices, but not the correct answer, during frame selection. These variants provide a reference with additional task information. Fixed queries produce one frame set for both QA tasks, whereas AR and TO queries select frames separately. Prompt templates and evaluation details appear in Appendix~\ref{sup:generic_query}.

\begin{table}[htbp]
\centering
\caption{Query-based samplers (\%). Details: Appendix~\ref{sup:generic_query}. Best: bold; second-best: underlined.}
\label{tab:fixed_query_main}
\small
\setlength{\tabcolsep}{3.5pt}
\begin{tabular}{@{} c ccc|cc|c @{}}
\toprule
\multirow{2}{*}{\textbf{Sampler}} & \multicolumn{3}{c|}{\textbf{SynFlash}} & \multicolumn{2}{c|}{\textbf{HIVAU}} & \textbf{V-MME} \\
\cmidrule(lr){2-4} \cmidrule(lr){5-6} \cmidrule(l){7-7}
& \textbf{R} & \textbf{AR} & \textbf{TO} & \textbf{ER} & \textbf{CR} & \textbf{Acc.} \\
\midrule
Uniform & 12.9 & 30.8 & 35.3 & \underline{60.6} & \underline{56.5} & \textbf{67.0} \\
\textbf{InfoShot} & \textbf{84.7} & \textbf{65.3} & \textbf{57.9} & \textbf{63.8} & \textbf{60.0} & \textbf{67.0} \\
\midrule
AKS (fixed) & 65.1 & 55.3 & \underline{51.2} & 27.8 & 31.5 & \underline{63.2} \\
GenS (fixed) & 15.3 & 31.5 & 34.0 & 29.9 & 37.4 & 48.2 \\
\midrule
AKS (AR) & \underline{71.7} & \underline{58.4} & --- & --- & --- & --- \\
AKS (TO) & 60.6 & --- & 49.7 & --- & --- & --- \\
GenS (AR) & 65.0 & 57.6 & --- & --- & --- & --- \\
GenS (TO) & 29.9 & --- & 37.6 & --- & --- & --- \\
\bottomrule
\end{tabular}
\end{table}

\noindent\textbf{Results.} Table~\ref{tab:fixed_query_main} shows that AKS and GenS with fixed queries obtain 65.1\% and 15.3\% SynFlash anomaly hit rate, respectively, compared with 84.7\% for InfoShot. Even with task questions, both samplers remain below InfoShot in average SynFlash AR and TO accuracy. On eight SynFlash videos, median selection time is 0.86 s for InfoShot, versus 4.90 s for GenS and 12.00 s for AKS with fixed queries on one B200 (Appendix~\ref{sup:timing}).

\section{Ablation Studies}
\label{sec:ablation}
\begin{table}[htbp]
\centering
\caption{Frame-role ablation (\%). Shots, frame budget, and downstream VLM are fixed. Best: bold; second-best: underlined.}
\label{tab:frame_roles}
\small
\setlength{\tabcolsep}{3.5pt}
\begin{tabular}{@{} l c|c|c @{}}
\toprule
\multirow{2}{*}{\textbf{Frame pair}} & \textbf{SynFlash} & \textbf{HIVAU} & \textbf{V-MME} \\
\cmidrule(lr){2-2} \cmidrule(lr){3-3} \cmidrule(l){4-4}
& \textbf{R} & \textbf{ER/CR} & \textbf{Acc.} \\
\midrule
Common+Common & 1.0 & 43.3/40.6 & \textbf{67.2} \\
Common+Random & \underline{8.7} & \underline{55.7/51.7} & \textbf{67.2} \\
Common+Unique & \textbf{84.7} & \textbf{63.8/60.0} & \underline{67.0} \\
\bottomrule
\end{tabular}
\end{table}

\subsection{Complementary Frame Roles}
\label{subsec:frame_roles}
With the shot partition, frame budget, and downstream VLM fixed, we replace the unique frame with a second common or random frame (Table~\ref{tab:frame_roles}).

Unique-frame selection raises SynFlash hit rate from 1.0\% (common) or 8.7\% (random) to 84.7\% and improves HIVAU coverage; Video-MME accuracy varies by 0.2 points across these three variants. This experiment does not measure the common frame's contribution to reasoning. Appendix~\ref{sup:attribution} reports hit attribution.
\subsection{Score-Weight Sensitivity}
\label{subsec:weight_sensitivity}
With $\alpha=0.5$, changing $\lambda$ among 0.5, 0.7, and 1.0 changes SynFlash hit rate by at most 0.43 points. The gain does not depend on the exact common-frame weight in these settings. At $\lambda=1.0$, using volatility alone ($\alpha=0$) reduces hit rate from 84.23\% to 66.25\%. These comparisons support combining atypicality and local change. Appendix Table~\ref{tab:score_weights} lists all tested settings.

\subsection{Feature-Extractor Comparison}
\label{subsec:feature_extractor}
InfoShot relies on feature-space similarity for shot boundary detection and within-shot keyframe selection. We ablate the feature extractor while keeping the segmentation and dual-keyframe selection fixed, under the SynFlash protocol with the same budget of 0.5 fps ($K=\lceil 0.5L\rceil$ for duration $L$). We compare HSV color histograms (global color distributions), ResNet-50 features (2048-d, ImageNet-pretrained), and SigLIP embeddings (language-aligned).

\begin{table}[htbp]
    \centering
    \caption{SynFlash feature ablation at 0.5 fps (\%). Best: bold; second-best: underlined.}

    \label{tab:feature_summary}
    \small
    \setlength{\tabcolsep}{2pt}
    \begin{tabular*}{\linewidth}{@{\extracolsep{\fill}}lllllll@{}}
        \toprule
        \textbf{Metric} & \textbf{Features} & \textbf{Camou.} & \textbf{Hi-Sat.} & \textbf{Curt.} & \textbf{PiP} & \textbf{Avg.} \\
        \midrule
        \multirow{3}{*}{\textbf{R}}
        & Histogram
        & 93.65 & 27.05 & 85.10 & 40.95 & 61.69 \\
        & SigLIP
        & \textbf{98.95} & \textbf{87.05} & \underline{87.35} & \textbf{74.80} & \textbf{87.04} \\
        & ResNet-50
        & \underline{98.35} & \textbf{87.05} & \textbf{88.45} & \underline{64.95} & \underline{84.70} \\
        \midrule
        \multirow{3}{*}{\textbf{AR}}
        & Histogram
        & 64.95 & \underline{35.85} & 71.15 & 39.55 & 52.88 \\
        & SigLIP
        & \textbf{69.60} & \textbf{70.60} & \textbf{74.10} & \textbf{52.40} & \textbf{66.68} \\
        & ResNet-50
        & \underline{69.15} & \textbf{70.60} & \underline{73.15} & \underline{48.35} & \underline{65.31} \\
        \midrule
        \multirow{3}{*}{\textbf{TO}}
        & Histogram
        & 58.80 & 38.20 & 58.45 & 44.45 & 49.98 \\
        & SigLIP
        & \underline{60.40} & \underline{56.75} & \textbf{61.95} & \textbf{55.45} & \textbf{58.64} \\
        & ResNet-50
        & \textbf{60.70} & \textbf{56.85} & \underline{61.10} & \underline{52.75} & \underline{57.85} \\
        \bottomrule
        \end{tabular*}
        \end{table}

\begin{figure}[!htbp]
\centering
\includegraphics[width=0.85\linewidth]{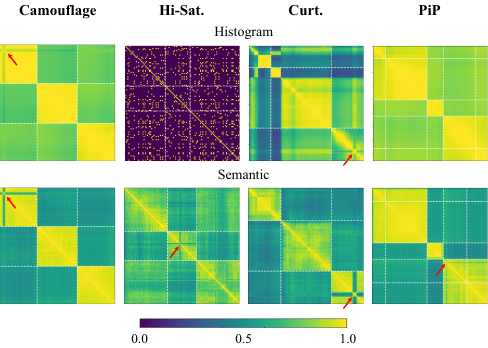}
\caption{Frame cosine similarities across all four SynFlash subsets. Rows compare histogram (upper) and semantic (bottom) features for the same videos; red arrows mark injected events.}
\label{fig:semantic_main}
\end{figure}

Semantic features are more robust to low-level photometric variation and better capture semantic content differences. Table~\ref{tab:feature_summary} shows higher average hit rate, AR, and TO under the same budget when using semantic features. SigLIP achieves the best overall performance, while ResNet-50 is close. We use ResNet-50 by default. Figure~\ref{fig:semantic_main} compares histogram and semantic features across all four SynFlash subsets; further analysis appears in Appendix~\ref{sup:feature_similarity}.

\begin{figure}[!htbp]
\centering
\includegraphics[width=0.65\linewidth]{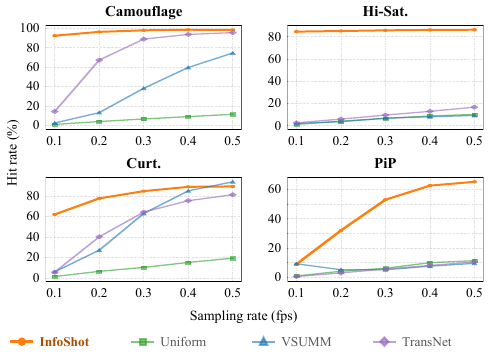}
\caption{SynFlash hit rate versus sampling rate. Full R/AR/TO results appear in Appendix Fig.~\ref{fig:budget_full}.}
\label{fig:fps_scaling}
\end{figure}

\subsection{Robustness to Frame Budget}
Figure~\ref{fig:fps_scaling} shows that InfoShot generally maintains higher hit rates across different anomaly patterns as the sampling rate decreases from 0.5 to 0.1 fps, indicating greater robustness to low frame budgets. Appendix~\ref{sup:budget} provides the full AR/TO curves and analysis.

\section{Conclusion}
InfoShot combines shot segmentation and complementary keyframes for task-agnostic video understanding under tight budgets. It improves anomaly coverage on SynFlash and HIVAU-70K while preserving Video-MME performance. On SynFlash, its average AR and TO accuracy exceeds GenS and AKS even with task questions. In TikTok's moderation pipeline, we also observed higher recall for evasive transient harmful content under tight compute constraints.

\noindent\textbf{Limitations and social impact.}
InfoShot uses a fixed two-frame allocation per shot, which may be suboptimal when shot lengths and event rates vary widely. Future work will explore adaptive budgets, uncertainty-driven selection, and lightweight motion cues. In moderation, retaining brief harmful content may reduce user exposure, but higher recall can increase false positives and review burden. Calibrated thresholds, human review, and evaluation across regions, languages, and content types remain necessary.

\bibliographystyle{IEEEbib}
\bibliography{main}

\clearpage
\appendix
\section{Additional Method Details}

These appendices accompany the main paper. Section~A expands the method in main-paper Sec.~\ref{sec:methodology}; Section~B complements the experiments and ablations in main-paper Secs.~\ref{sec:experiments} and~\ref{sec:ablation}; Section~C details the SynFlash protocol; and Section~D presents qualitative successes and failure cases. Main-paper references are identified explicitly below.

\subsection{Information-Theoretic Analysis}
\label{sup:info_maximization}

Let $G$ denote dominant shot content and temporal structure, including the partition and background parameters, and let $E$ denote sparse-event states, content, and locations. The binary sequences $S_{1:T}$ and $F_{1:T}$ describe boundaries and event occurrence, respectively; they do not alone specify the semantic content carried by a frame. We therefore use $(G,E)$ as the information target.

\noindent\textbf{Background and event model.}
In main-paper Eq.~\ref{eq:stochastic_model}, $b_t=\mu_{m(t)}+\epsilon_t$ is a latent background process. Within a segment, $\epsilon_t$ has locally stationary Gaussian marginals and weak temporal dependence. The event component $\eta_t$ has an event-dependent distribution and need not follow the background distribution. Thus the observation $c_t=(1-f_t)b_t+f_t\eta_t$ may depart abruptly from the background. Local stationarity concerns $b_t$, not the complete event-contaminated sequence. The model does not impose a random walk on the background or assume that every event is a large feature outlier.

\noindent\textbf{Information decomposition.}
For a fixed candidate index set $\mathcal K$, assuming the relevant mutual informations exist, the chain rule gives
\begin{equation}
\begin{aligned}
I(C_{\mathcal K};G,E)
&=I(C_{\mathcal K};G)\\
&\quad+I(C_{\mathcal K};E\mid G).
\end{aligned}
\end{equation}
The first term concerns representative content and structure. The second concerns event evidence beyond that context. For $i\notin\mathcal K$, the information added by another frame is
\begin{equation}
\begin{aligned}
&I(C_{\mathcal K\cup\{i\}};G,E)-I(C_{\mathcal K};G,E)\\
&\qquad=I(c_i;G,E\mid C_{\mathcal K}).
\end{aligned}
\end{equation}
This conditional quantity distinguishes additional evidence from repeated observations of the same content.

For a simple illustration, let $G$ and $E$ be independent binary variables, and consider $c_1=c_2=G$ and $c_3=E$. Then
\begin{equation}
\begin{aligned}
I(c_1,c_2;G,E)&=H(G),\\
I(c_1,c_3;G,E)&=H(G)+H(E).
\end{aligned}
\end{equation}
When $H(E)>0$, the complementary pair contains more information than the repeated representative pair. This discrete illustration explains the design motivation; it is not a model of the measured videos or a proof that the proposed scores optimize mutual information. A high-entropy observation independent of both latent components also need not provide useful information.

The identities above concern fixed index sets. InfoShot selects indices from the observed feature sequence, so guarantees for its adaptive selection would require further analysis. We neither estimate mutual information nor claim global optimality of the greedy partition or the common--unique pair.

\subsection{Rationale for Dual-Keyframe Selection}
\label{sup:dualframe_rationale}
For a segment $m$, let $G_m$ describe its dominant content and $E_m$ its event state and content. The local counterpart of the main-paper decomposition is
\begin{equation}
\begin{aligned}
I(c_i,c_j;G_m,E_m)
&=I(c_i,c_j;G_m)\\
&\quad+I(c_i,c_j;E_m\mid G_m).
\end{aligned}
\label{eq:withinshot_chain}
\end{equation}
This motivates two intended roles under a two-frame allocation: retaining representative context and adding evidence about deviations from that context. It does not establish that two frames are sufficient for every segment.

\noindent\textbf{Representative context.}
Write $z_i=c_i/\|c_i\|_2$. Since $A_{ij}=z_i^\top z_j$, the typicality score satisfies
\begin{equation}
1-g_i=\frac{1}{2(L-1)}\sum_{j\in\mathcal S_m,\,j\ne i}\|z_i-z_j\|_2^2.
\end{equation}
For $L>1$, maximizing $g_i$ therefore selects a medoid under squared Euclidean distance between normalized features. The common-frame rule additionally penalizes local volatility when $\lambda<1$, favoring a representative that is locally stable.

\noindent\textbf{Deviation evidence.}
The volatility score similarly satisfies
\begin{equation}
v_i=\frac{1}{2|\mathcal N_i|}\sum_{j\in\mathcal N_i}\|z_i-z_j\|_2^2.
\end{equation}
Atypicality measures departure from the segment as a whole, while volatility measures local change. Their normalized combination targets deviations that a representative frame may omit. These cues can also respond to motion or noise; neither is an estimator of semantic information. The utility of the unique-frame rule is evaluated by the replacement experiment in main-paper Table~\ref{tab:frame_roles}, which does not separately isolate the common frame's contribution to reasoning. The rules enforce $t_{\mathrm{uni}}\ne t_{\mathrm{com}}$.

\subsection{Windowed Contrast and Mean-Change Tests}
\label{sup:boundary_score}
For a candidate boundary, let the clipped left and right windows contain $n$ and $m$ normalized features, respectively. Define
\begin{equation}
\begin{aligned}
U_-&=\frac{\sum_{i\ne j\in\mathcal W^-}A_{ij}}{n(n-1)},\\
U_+&=\frac{\sum_{i\ne j\in\mathcal W^+}A_{ij}}{m(m-1)},\\
V&=\frac{\sum_{i\in\mathcal W^-,j\in\mathcal W^+}A_{ij}}{nm}.
\end{aligned}
\end{equation}
For $n,m>1$, the implemented contrast is $B_t=(U_-+U_+)/2-V$. Within-window averages exclude diagonal self-similarities. For a singleton window, the implementation sets its within-window average to zero; the formulas with $n-1$ or $m-1$ below apply only when both windows have at least two frames.

\noindent\textbf{Two-sample interpretation.}
The expression $2B_t=U_-+U_+-2V$ has the algebraic form of the two-sample MMD U-statistic under the normalized linear kernel~\cite{gretton2012kernel}. With independent samples, it is an unbiased estimator of the corresponding population squared mean difference. Video features may be temporally dependent; that statistical property does not follow from the algebraic identity alone. A linear kernel also need not distinguish distributions with identical feature means.

Let $\bar z_-$ and $\bar z_+$ be the two window means. Unit feature norms give the exact expansion
\begin{equation}
\begin{aligned}
B_t={}&\tfrac12\|\bar z_- -\bar z_+\|_2^2\\
&-\tfrac12\left[\frac{1-\|\bar z_-\|_2^2}{n-1}
+\frac{1-\|\bar z_+\|_2^2}{m-1}\right].
\end{aligned}
\label{eq:contrast_mean_difference}
\end{equation}
This expresses a mean-change contrast with finite-window corrections. The sample statistic can be negative.

\noindent\textbf{Relation to a Gaussian GLRT.}
As a working approximation, consider independent observations with known common covariance $\sigma^2 I$. Under $H_0$, the two windows share an unknown mean; under $H_1$, they have separate unknown means. Maximizing both likelihoods gives
\begin{equation}
\begin{aligned}
\log\Lambda_t
&=\frac{\mathrm{RSS}_0-\mathrm{RSS}_1}{2\sigma^2}\\
&=\frac{nm}{2\sigma^2(n+m)}\|\bar z_- -\bar z_+\|_2^2,
\end{aligned}
\end{equation}
where RSS denotes the fitted residual sum of squares. This Gaussian model is an approximation to local features, not an exact distribution for unit-normalized vectors. It connects mean changes to likelihood-ratio testing, but the implemented $B_t$ also contains the corrections in Eq.~\ref{eq:contrast_mean_difference}, and its clipped window sizes can vary. We therefore describe the method as windowed contrast rather than identify its score with an exact GLRT. No test threshold, false-positive guarantee, or optimal partition follows from this connection; the algorithm ranks boundaries under its frame budget.

\Needspace{38\baselineskip}
\subsection{InfoShot Algorithm}
\label{sup:pseudocode}

\begin{algorithm}[H]
\small
\algrenewcommand\algorithmicindent{1em}
\caption{InfoShot}
\label{alg:video_sampling}
\textbf{Input:} Feature sequence $C=\{c_1,\dots,c_T\}$, frame budget $2\le K\le T$, weights $\lambda,\alpha$. \\
\textbf{Output:} Sampled keyframe indices $\mathcal{K}$.
\begin{algorithmic}[1]
\State Compute affinity $A_{ij}=\cos(c_i,c_j)$.
\State $M \leftarrow \lfloor K/2 \rfloor$ ~~~\text{(two frames per shot)}
\State \textbf{Initialize:} partition $\mathcal{P}\leftarrow\{[1,T]\}$, $\mathcal{K}\leftarrow\emptyset$.
\vspace{1mm}
\State \textbf{Phase 1: Greedy shot segmentation}
\While{$|\mathcal{P}|<M$}
    \State For each segment $[a,b]\in\mathcal{P}$ with $\mathcal T_{a,b}\ne\emptyset$, compute using main-paper Eq.~\ref{eq:boundary_score}:
    \Statex \hspace{\algorithmicindent}$t^\ast_{[a,b]} \leftarrow \arg\max_{t\in\mathcal T_{a,b}} B_t$.
    \If{no segment admits a valid split} \State \textbf{break} \EndIf
    \State Select the best valid split:
    \Statex \hspace{\algorithmicindent}$\begin{aligned}
      ([a^\star,b^\star],t^\star)&\leftarrow\underset{[a,b]\in\mathcal{P}:\,\mathcal T_{a,b}\ne\emptyset}{\arg\max}\, B_{t^\ast_{[a,b]}}.
    \end{aligned}$
    \State Replace $[a^\star,b^\star]$ in $\mathcal{P}$ by $[a^\star,t^\star-1]$ and $[t^\star,b^\star]$.
\EndWhile
\vspace{1mm}
\State \textbf{Phase 2: Dual-keyframe selection}
\For{each shot segment $\mathcal{S}_m=[a,b]\in\mathcal{P}$}
    \If{$|\mathcal{S}_m|=1$}
        \State $\mathcal{K}\leftarrow \mathcal{K}\cup\{a\}$; \textbf{continue}
    \EndIf
    \State For all $i\in\mathcal{S}_m$, compute
    \Statex \hspace{2em}$g_i \leftarrow \mathbb{E}_{j\in\mathcal{S}_m\setminus\{i\}}[A_{ij}]$,
    \Statex \hspace{2em}$v_i \leftarrow 1-\mathbb{E}_{j\in\mathcal{N}_i}[A_{ij}]$.
    \State Normalize $g_i,v_i$ within $\mathcal{S}_m$ to obtain $\hat g_i,\hat v_i\in[0,1]$.
    \State $t_{\mathrm{com}}\leftarrow \arg\max_{i\in\mathcal{S}_m}\big(\lambda \hat g_i-(1-\lambda)\hat v_i\big)$.
    \State $\displaystyle t_{\mathrm{uni}}\leftarrow \underset{i\in\mathcal{S}_m\setminus\{t_{\mathrm{com}}\}}{\arg\max}$
    \Statex \hspace{2em}$\big(\alpha(1-\hat g_i)+(1-\alpha)\hat v_i\big)$.
    \State $\mathcal{K}\leftarrow \mathcal{K}\cup\{t_{\mathrm{com}},t_{\mathrm{uni}}\}$.
\EndFor
\State \textbf{return} $\mathcal{K}$
\end{algorithmic}
\end{algorithm}

\noindent\textbf{Implementation notes.}
Algorithm~\ref{alg:video_sampling} describes the two-keyframe core for $K\ge2$. The set $\mathcal T_{a,b}$ contains valid boundary indices $t\in\{a+1,\ldots,b\}$ satisfying the window and shot-length constraints in the main-paper implementation. Empty candidate sets are skipped. The boundary index denotes the first frame of the right segment, consistent with the windows $\mathcal W_t^-$ and $\mathcal W_t^+$ in main-paper Sec.~\ref{subsec:shot_seg}.

With at most $M=\lfloor K/2\rfloor$ shots and at most two selected frames per shot, $|\mathcal K|\le2M\le K$. Thus $K$ is a budget cap; an odd $K$ need not be fully used by the two-frame rule. Singleton shots contribute one frame. This pseudocode does not specify a separate one-frame-budget policy. We enforce $t_{\mathrm{uni}}\ne t_{\mathrm{com}}$ and sort selected indices chronologically before VLM inference.

\subsection{Computational Considerations}
\label{sup:complexity}
Sampling uses a visual encoder and feature-space operations, with no downstream VLM calls during frame selection. This subsection distinguishes arithmetic simplifications from measured runtime.

Let $z_i=c_i/\|c_i\|_2$ and $h_S=\sum_{j\in S}z_j$. For a shot $S$ of length $L>1$, the typicality and volatility scores can be written as
\begin{equation}
\begin{aligned}
g_i&=\frac{z_i^\top h_S-1}{L-1},\\
v_i&=1-\frac{z_i^\top h_{\mathcal{N}_i}}{|\mathcal{N}_i|}.
\end{aligned}
\end{equation}
Shot sums and temporal prefix sums avoid materializing all pairwise affinities for these scores. For a fixed partition, evaluating all common/unique scores takes $O(Td)$ arithmetic for $d$-dimensional features. Likewise, each within-window or cross-window average in main-paper Eq.~\ref{eq:boundary_score} can be evaluated from window sums: a cross average is the inner product of the two mean feature vectors. This makes one sweep of boundary scores linear in $Td$.

The complete sampler also performs greedy split selection. Writing its search and score-recomputation cost as $C_{\mathrm{split}}(T,M)$, an aggregate-sum implementation has cost $O(TC_\phi+Td+C_{\mathrm{split}}(T,M))$, excluding video decoding, where $C_\phi$ is the per-frame encoder cost and $M$ is the target number of shots. The full greedy procedure is not established to be linear merely by the score identities above. The blockwise affinity implementation described in the main paper instead materializes similarities within blocks of at most 600 frames; its affinity arithmetic is $O(d\sum_b n_b^2)$ for block lengths $n_b$. These descriptions are implementation alternatives, not wall-clock measurements. We do not infer a speedup over other samplers from this analysis. All compared frame sets subsequently enter the same VLM under the specified frame budget.

\section{Additional Experiments}

\subsection{Hit Attribution}
\label{sup:attribution}
As a diagnostic accompanying main-paper Table~\ref{tab:frame_roles}, aggregate attribution over SynFlash reports 0.5\% of videos hit by common frames only, 84.2\% by unique frames only, and 15.3\% missed by the selected set. At the reported one-decimal precision, the exclusive-hit percentages sum to 84.7\%, matching the reported default hit rate of 84.70\%. These observations describe where selected evidence occurs; the controlled replacement experiment in the main paper provides the comparison against a second representative or random frame.

\subsection{Score-Weight Sensitivity}
\label{sup:parameters}
Table~\ref{tab:score_weights} varies the weights in main-paper Eq.~\ref{eq:uni_com}. The default $(\lambda,\alpha)=(0.7,0.5)$ gives the highest reported hit rate among these settings. Holding $\alpha=0.5$, replacing $\lambda=0.7$ by 0.5 or 1.0 changes hit rate from 84.66 to 84.23. The volatility-only unique score ($\alpha=0$) paired with $\lambda=1.0$ drops to 66.25. Because some rows vary both weights, the table should be interpreted as a sensitivity check over the listed configurations rather than a full factorial analysis.

\begin{table}[t]
\centering
\caption{SynFlash hit rate (\%) for the score-weight configurations evaluated. The first row is the default. Best: bold; second-best: underlined.}
\label{tab:score_weights}
\begin{tabular}{@{}ccc@{}}
\toprule
$\boldsymbol{\lambda}$ & $\boldsymbol{\alpha}$ & \textbf{SynFlash R} \\
\midrule
0.7 & 0.5 & \textbf{84.66} \\
0.5 & 0.5 & \underline{84.23} \\
1.0 & 0.5 & \underline{84.23} \\
0.0 & 1.0 & 83.63 \\
0.7 & 0.25 & 77.34 \\
1.0 & 0.0 & 66.25 \\
\bottomrule
\end{tabular}
\end{table}

\Needspace{8\baselineskip}
\subsection{Feature-Space Coverage Analysis}
\label{sup:distortion}

\noindent\textbf{Feature-space coverage.}
Beyond anomaly hit rate, we also evaluate how well the sampled frames cover a fixed feature space under the same budget $K$. Following main-paper Sec.~\ref{sec:methodology}, given $\mathcal{V}=\{x_t\}_{t=1}^{T}$ and features $c_t=\phi(x_t)$, a sampler outputs indices $\mathcal{K}\subset\{1,\dots,T\}$ with $|\mathcal{K}|\le K$ and the sampled set $C_{\mathcal{K}}=\{c_t\}_{t\in\mathcal{K}}$. Retaining both representative context and rare deviations is intended to improve coverage of the full sequence $C_{1:T}=\{c_t\}_{t=1}^{T}$.

We measure this via set distortion under cosine distance. We first normalize features, and define $A_{ij}=\cos(c_i,c_j)$ as in Sec.~\ref{subsec:shot_seg}. The distortion of a sampled set is
\begin{equation}
\begin{aligned}
\mathrm{Dist}(\mathcal{K})
&=\frac{1}{T}\sum_{t=1}^{T}\min_{k\in\mathcal{K}}\bigl(1-A_{tk}\bigr)\\
&=1-\frac{1}{T}\sum_{t=1}^{T}\max_{k\in\mathcal{K}} A_{tk}.
\end{aligned}
\end{equation}
Intuitively, for each full frame $t$ we find the most similar sampled frame $k\in\mathcal{K}$ and average the residual $1-A_{tk}$. Lower $\mathrm{Dist}(\mathcal{K})$ indicates better feature-space coverage under the same budget. Main-paper Fig.~\ref{fig:coverage_main} reports $\mathrm{Dist}(\mathcal{K})$ at 0.5 fps on SynFlash using the existing values at their reported three-decimal precision. InfoShot gives the lowest distortion on Camouflage, High-saturation, and Curtain. VSUMM has slightly lower distortion on PiP (0.075 versus 0.076 for InfoShot). This exception illustrates that average feature-space coverage and transient-event hit rate need not rank samplers identically. The metric describes coverage in the evaluated feature space; it is not an estimate of mutual information or an encoder-independent measure of retained information. The experiment also does not isolate the contribution of the common frame.

\subsection{Feature Extractor}
\label{sup:feature_extractor}
\label{sup:feature_similarity}
\noindent\textbf{Setup.}
InfoShot relies on feature-space similarity for shot boundary detection and within-shot keyframe selection. We ablate the feature extractor while keeping the segmentation and dual-keyframe selection fixed, under the SynFlash protocol with the same budget of 0.5 fps ($K=\lceil 0.5L\rceil$ for duration $L$). We compare HSV color histograms (global color distributions), ResNet-50 features (2048-d, ImageNet-pretrained), and SigLIP embeddings (language-aligned).

\noindent\textbf{Semantic features.}
In main-paper Fig.~\ref{fig:semantic_main}, we keep the segmentation and dual-keyframe selection unchanged and only replace the frame feature representation. It shows that histogram features are unreliable on SynFlash: in Hi-Sat., rapid background flickering changes global color statistics and produces speckle-like similarity patterns; in PiP, the small insertion may not shift the global histogram enough to stand out. Histogram cues are also weak in Camouflage (color matching keeps histograms similar) and in Curtain (transition-induced color shifts make normal scene changes resemble injected events). As a result, histogram similarity is not a reliable proxy for the semantic evidence needed by AR and TO. Additional case studies are provided in Appendix~\ref{sup:case_study}.

Semantic features are more robust to low-level photometric variation and better capture semantic content differences. Main-paper Table~\ref{tab:feature_summary} reports the feature-extractor comparison. Both learned representations improve the average R, AR, and TO over histograms. SigLIP improves average hit rate by 2.34 percentage points over ResNet-50, although its Curtain hit rate is lower (87.35\% versus 88.45\%). SigLIP achieves the best overall performance, while ResNet-50 is close. We keep ResNet-50 as the default. This comparison changes the feature representation; it does not establish the absence of overlap between encoder pretraining data and SynFlash source images, and it does not measure extraction latency.

\subsection{Robustness to Frame Budget}
\label{sup:budget}

\begin{figure}[htbp]
    \centering
    \includegraphics[width=0.98\textwidth]{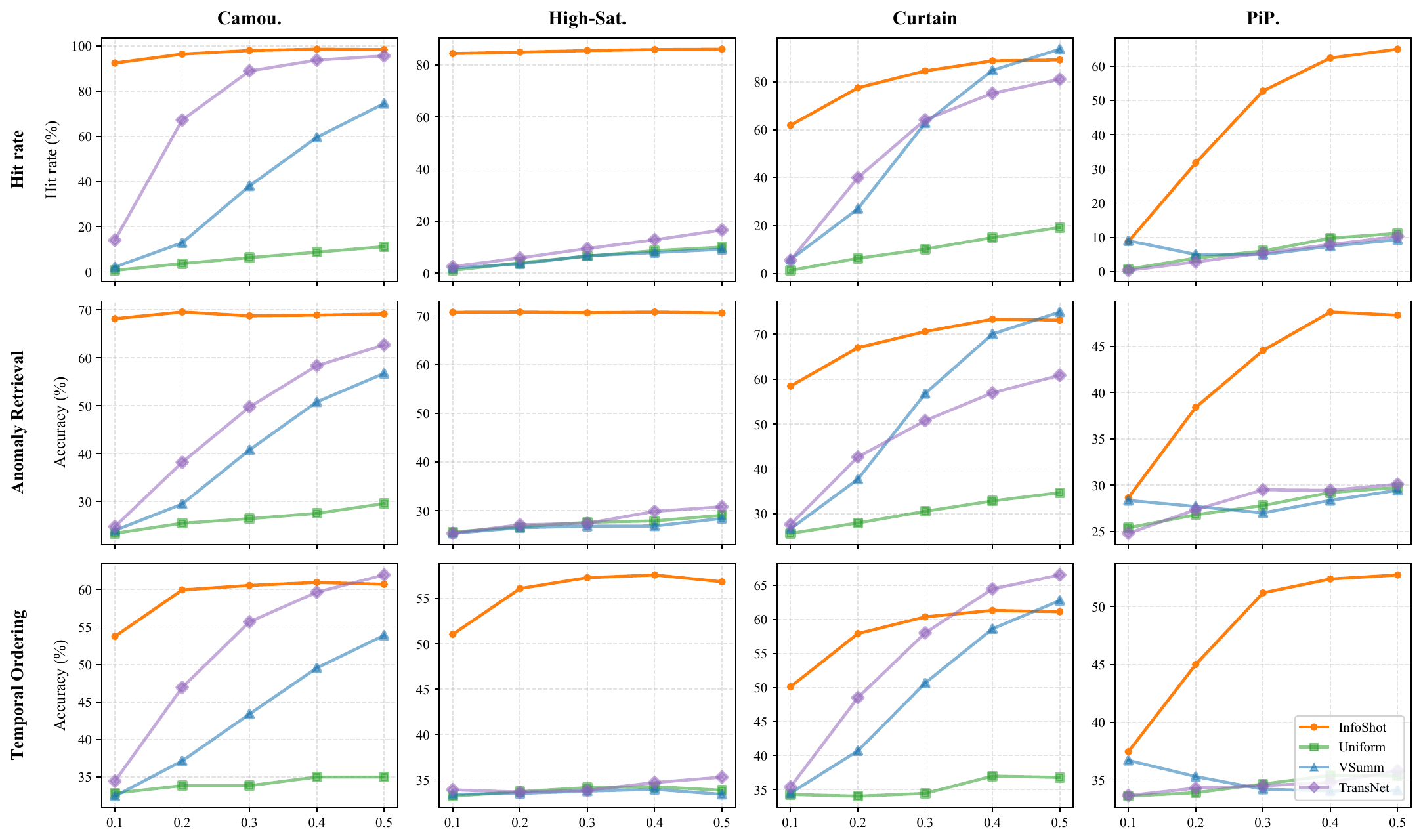}
    \caption{Full SynFlash frame-budget results: anomaly hit rate, anomaly retrieval (AR), and temporal ordering (TO) over sampling rates from 0.1 to 0.5 fps. The main paper retains the hit-rate row.}
    \label{fig:budget_full}
\end{figure}

We evaluate robustness under sampling rates from 0.1 to 0.5 fps. Figure~\ref{fig:budget_full} reports anomaly hit rate, anomaly retrieval (AR), and temporal ordering (TO) across the four SynFlash subsets. The curves characterize the tradeoff between sampling budget, evidence coverage, and downstream accuracy, including the sparse 0.1 fps regime.

From an information perspective, extremely sparse sampling makes the selected set a hard bottleneck: if the anomaly evidence is not sampled, VQA becomes under-determined regardless of the VLM. In Fig.~\ref{fig:budget_full}, uniform sampling exhibits very low hit rate at 0.1 fps and VSUMM is limited by redundancy under tight budgets. Boundary-only approaches, TransNet, improve hit rate for subsets with clear transitions but remain weak on localized patterns (e.g., Hi-Sat. and PiP). In contrast, InfoShot mitigates this bottleneck by allocating budget at the shot level (reducing wasted samples in long static content) and explicitly selecting, within each shot, a representative frame and a high-deviation frame. This combination increases the probability that the sampled set retains both global context and short, localized evidence, which explains the stable gains in recall and the corresponding improvements in VQA at lower fps settings.

\FloatBarrier
\subsection{Comparison with Query-Based Samplers}
\label{sup:generic_query}
We compare AKS-SeViLA~\cite{tang2025adaptive,yu2023self} and GenS~\cite{yao2025generative} with and without the downstream question. In the fixed setting, both methods use the same query for all examples:
\begin{quote}
A compact visual summary of the video covering scenes, objects, actions, temporal changes, and salient moments.
\end{quote}
In the AR and TO settings, the sampler receives the corresponding question and answer choices, but not the correct answer. Each setting selects its own frames, so AR and TO have separate hit-rate values. The fixed setting uses one frame set for both tasks. The two methods receive the same query content but keep their own prompt templates and frame-selection rules.

\noindent\textbf{Prompt templates.}
GenS receives numbered candidate images, followed by the text below. Here \texttt{query} is either the fixed query above or the task question followed by its answer choices.
\begin{quote}\small
Question: \texttt{\{query\}}\\
Please identify the video frames most relevant to the given question (with options) and provide their timestamps in seconds along with a relevance score. The score should be on a scale from 1 to 5, where higher scores indicate greater relevance. Return the output strictly in the following JSON format: \texttt{\{"timestamp": score, ...\}}. If no relevant frames are found, return \texttt{[None]}.
\end{quote}
We retain the template wording. In the implementation used here, returned keys are interpreted as frame indices within each window. The fixed parser accepts explicit index--score pairs; unusable outputs are retained as misses.

AKS uses the following original template, including its wording:
\begin{quote}\small
Question: \texttt{\{question\}} Candidate: \texttt{\{options\}}. Is this a good frame can answer the question?
\end{quote}
For fixed queries, \texttt{question} is the generic query and \texttt{options} is empty. Otherwise, they contain the task question and concatenated answer choices. The original text processor keeps at most 50 words, with a 77-token model limit; 292 of the 16,000 task questions are truncated. Neither template receives the correct answer.

\noindent\textbf{Sampling and QA.}
The new SynFlash runs cover all 8,000 videos at a nominal budget of 0.5 fps, using 10 fps candidate frames. GenS uses its Qwen2.5-VL-3B selector; AKS uses the SeViLA scorer with BLIP-2/Flan-T5-XL and the AKS selection algorithm. Both use Qwen2.5-VL-32B-Instruct for downstream QA with greedy decoding and at most eight new tokens. We keep the selected frame counts, including cases where AKS returns fewer than the target budget. Its fixed setting selects 66,264 frames against a total target of 87,286. All planned examples remain in the denominator: missing or invalid selections count as misses, and missing answers count as incorrect.

All SynFlash hit rates use the shared tolerance-window protocol defined in main-paper Sec.~\ref{subsec:benchmark_design}. Original InfoShot and task-free baseline scores are retained from the paper, rather than rerun under the new decoding pipeline. The new AKS SynFlash measurements replace the earlier fixed-query hit rate of 38.1\%; AKS HIVAU and Video-MME scores are carried over from the existing evaluation, rather than rerun in this round.

\noindent\textbf{Results by subset.}
Table~\ref{tab:query_subsets} reports the new SynFlash results. Giving the question helps GenS on both QA tasks and AKS on AR, but AKS TO decreases from 51.16\% to 49.71\%. The average AR and TO scores remain below the reported InfoShot scores. This does not hold for every subset: GenS with the AR question reaches 75.20\% AR on Curtain, above InfoShot's 73.15\%, and AKS reaches 51.95\% on PiP, above 48.35\%.

\begin{table}[!thbp]
\centering
\caption{SynFlash results by sampler and query setting (\%). R is hit rate under the stated tolerance windows. Fixed uses a generic query; AR and TO supply the corresponding question and options. Best: bold; second-best: underlined.}
\label{tab:query_subsets}
\small
\setlength{\tabcolsep}{2pt}
\begin{tabular*}{\linewidth}{@{\extracolsep{\fill}}lllllll@{}}
\toprule
\textbf{Metric} & \textbf{Sampler} & \textbf{Camou.} & \textbf{Hi-Sat.} & \textbf{Curt.} & \textbf{PiP} & \textbf{Avg.} \\
\midrule
R & \textbf{InfoShot} & \textbf{98.35} & \textbf{87.05} & \underline{88.45} & \underline{64.95} & \textbf{84.70} \\
R & AKS (fixed) & 72.85 & 49.95 & 80.90 & 56.75 & 65.11 \\
R & AKS (AR) & \underline{82.15} & 52.80 & 85.05 & \textbf{66.85} & \underline{71.71} \\
R & AKS (TO) & 77.40 & 38.50 & 75.75 & 50.60 & 60.56 \\
R & GenS (fixed) & 16.65 & 13.85 & 17.95 & 12.75 & 15.30 \\
R & GenS (AR) & 80.95 & \underline{67.90} & \textbf{93.25} & 17.90 & 65.00 \\
R & GenS (TO) & 35.00 & 28.30 & 41.65 & 14.55 & 29.88 \\
\midrule
AR & \textbf{InfoShot} & \textbf{69.15} & \textbf{70.60} & \underline{73.15} & \underline{48.35} & \textbf{65.31} \\
AR & AKS (fixed) & 57.30 & 49.90 & 67.20 & 46.90 & 55.33 \\
AR & AKS (AR) & 61.10 & 51.25 & 69.20 & \textbf{51.95} & \underline{58.38} \\
AR & GenS (fixed) & 32.20 & 28.65 & 34.80 & 30.15 & 31.45 \\
AR & GenS (AR) & \underline{62.50} & \underline{58.50} & \textbf{75.20} & 34.00 & 57.55 \\
\midrule
TO & \textbf{InfoShot} & \textbf{60.70} & \textbf{56.85} & \textbf{61.10} & \textbf{52.75} & \textbf{57.85} \\
TO & AKS (fixed) & 51.80 & \underline{46.35} & \underline{56.25} & \underline{50.25} & \underline{51.16} \\
TO & AKS (TO) & \underline{52.45} & 43.25 & 54.95 & 48.20 & 49.71 \\
TO & GenS (fixed) & 34.70 & 33.10 & 33.90 & 34.40 & 34.03 \\
TO & GenS (TO) & 38.45 & 35.75 & 39.00 & 37.25 & 37.61 \\
\bottomrule
\end{tabular*}
\end{table}

\noindent\textbf{Other benchmarks.}
GenS with the fixed query obtains 29.9\% ER and 37.4\% CR on HIVAU at 0.1 fps, over 628 videos and 1,382 events. On Video-MME short and medium videos at 0.25 fps, fixed and question-based selection yield 48.2\% and 54.1\% accuracy. These use Qwen2.5-VL-7B-Instruct, no subtitles, and the original QA script with a 16-token generation limit. We use the same answer parsing and accuracy calculation as Table~\ref{tab:videomme_mini} in the main paper. Main-paper Table~\ref{tab:fixed_query_main} reports the GenS fixed-query result.

\FloatBarrier
\subsection{Frame-Selection Time}
\label{sup:timing}
Table~\ref{tab:selection_time} reports median selection time on eight SynFlash videos, two per injection pattern, at a target budget of 0.5 fps. GPU methods use one B200; Uniform and VSUMM run on CPU. After one warm-up run, we take the median of three runs per video and report the median across videos. Timing includes feature extraction and selection, but excludes video decoding, model loading, and downstream QA. Candidate grids and actual selected-frame counts follow each method's implementation. TransNet uses its PyTorch port, checked against the original TensorFlow model.

\begin{table}[htbp]
\centering
\caption{Median frame-selection time per video on eight SynFlash videos. Fixed, AR, and TO follow the query settings above.}
\label{tab:selection_time}
\small
\begin{tabular}{@{}lcc@{}}
\toprule
\textbf{Sampler} & \textbf{Query} & \textbf{Time (s)} \\
\midrule
Uniform & --- & $<0.001$ \\
VSUMM & --- & 0.19 \\
TransNet (PyTorch) & --- & 0.10 \\
InfoShot & --- & 0.86 \\
\midrule
GenS & Fixed & 4.90 \\
GenS & AR & 3.27 \\
GenS & TO & 3.96 \\
\midrule
AKS & Fixed & 12.00 \\
AKS & AR & 11.94 \\
AKS & TO & 12.03 \\
\bottomrule
\end{tabular}
\end{table}

On these videos, InfoShot is faster than GenS and AKS, while Uniform, VSUMM, and TransNet are faster than InfoShot. These measurements concern selection time for this sample, rather than end-to-end runtime.

\subsection{Motion-Heavy Video Check}
\label{sup:diving}
We additionally examine 500 Diving48 test videos~\cite{li2018resound} using a sampling budget of 2 fps. Diving48 supplies video-level action labels rather than transient-event intervals. We construct three multiple-choice questions per video concerning takeoff group, coarse flight movement, and entry position. Table~\ref{tab:diving_vqa} reports this Diving48-style VQA check, which differs from the original 48-class action-recognition evaluation.

\begin{table}[htbp]
\centering
\caption{Diving48-style VQA accuracy (\%) on 500 videos at 2 fps. Best: bold; second-best: underlined.}
\label{tab:diving_vqa}
\small
\begin{tabular*}{\linewidth}{@{\extracolsep{\fill}}lllll@{}}
\toprule
\textbf{Sampler} & \textbf{Takeoff} & \textbf{Flight} & \textbf{Entry} & \textbf{Overall} \\
\midrule
Uniform & \underline{19.2} & \textbf{43.2} & \underline{26.0} & \underline{29.5} \\
InfoShot & \textbf{21.4} & \textbf{43.2} & \textbf{28.4} & \textbf{31.0} \\
\bottomrule
\end{tabular*}
\end{table}

InfoShot improves overall accuracy from 29.5 to 31.0, with identical flight accuracy. The gain is smaller than the transient-event gains on SynFlash. This limited comparison indicates that performance does not collapse on these motion-heavy clips; it does not establish superiority over motion-aware samplers or statistical significance. Sustained high volatility remains a limitation of a fixed two-frame allocation per shot.

\Needspace{8\baselineskip}
\section{SynFlash Construction and Evaluation Protocol}
\label{sup:synflash_protocol}

This section provides additional implementation details for SynFlash beyond the main paper, including the pattern-specific injection procedure, annotation and multiple-choice construction, and downstream VLM evaluation.

\subsection{Injection Procedure}

Each SynFlash sample is generated by selecting one of the three constituent clips as the target clip for anomaly insertion and then sampling a valid insertion position within that clip. The selected clip index also determines the coarse temporal label used in evaluation. The anomaly source image is sampled from a valid ImageNet-1k class folder, and its class name is recorded as the anomaly label.

We implement four injection patterns with different spatial and temporal characteristics. In \emph{Camouflage}, the anomaly is inserted as a short full-frame transient whose color statistics are matched to the local background in CIELAB space, followed by a mild temporal zoom and smooth blending. In \emph{Curtain}, the anomaly is introduced through a column-wise wipe transition, where the valid image region is partitioned into vertical columns and the anomaly progressively appears and disappears through wipe-in and wipe-out stages. In \emph{Picture-in-picture (PiP)}, the anomaly is inserted only within a local patch at a random spatial location. The patch height is set to approximately 35\% of the valid image height, while the patch width is determined by the original aspect ratio of the injected image. In \emph{High-saturation}, the original frame is embedded into a smaller inner canvas, while the outer border is filled with rapidly changing saturated colors; the anomaly is then injected into the inner region for a short duration, creating strong distractive flicker unrelated to the true anomaly.

All videos are normalized to 24 fps and $640\times480$ resolution. Durations are pattern-specific. For Camouflage and PiP, the anomaly typically lasts 5--7 frames (about 0.21--0.29 seconds). For the Curtain subset, the duration is determined by the wipe-in, hold, and wipe-out stages. For the High-saturation subset, the anomaly lasts about 0.1 seconds, while the border colors change continuously throughout the video.

\subsection{Annotation, Multiple-Choice Construction, and Evaluation}

For each generated video, we store the anomaly label and the insertion timestamp metadata. Anomaly hit rate follows the shared tolerance-window protocol in main-paper Sec.~\ref{subsec:benchmark_design}. These metadata are used to automatically generate the downstream question-answer pairs. For AR, the correct answer is the ImageNet label of the injected anomaly. For TO, the answer is derived from the recorded insertion metadata.

To reduce ambiguity in answer matching, we convert each question into a multiple-choice format. For AR, we generate three distractors using a Qwen-family text-only large language model conditioned on the question and ground-truth answer, with prompts that require distractors from the same superclass or type as the correct answer, as described in the main-paper AR protocol. For TO, the options are drawn from a fixed label set. The correct answer and distractors are then shuffled and mapped to answer letters.

Given a sampler output, we collect the selected frames, sort them by frame index, and provide them to the VLM as an ordered image sequence. For each question, we construct a prompt containing the question text and candidate options. During inference, the VLM generates both an answer label and a short textual reason. As in the main paper, we use Qwen2.5-VL with deterministic decoding (temperature 0), identical question prompts across samplers, and frames in temporal order. The task-agnostic samplers do not receive the downstream question or answer choices; query-based settings are specified in Sec.~\ref{sup:generic_query}. For quantitative scoring, we extract the predicted option label from the generated response and compare it against the stored ground-truth label. The generated reasons are retained for qualitative analysis. Evaluation is parallelized across multiple GPUs by partitioning the dataset into disjoint subsets and merging the prediction files after all worker processes finish.

\section{SynFlash Case Studies}
\label{sup:case_study}
In this section, we present additional SynFlash case studies covering all four injection patterns, and analyze how the sampled evidence influences downstream VLM performance on both anomaly retrieval (AR) and temporal ordering (TO).

\subsection{Camouflage}

\begin{figure}[htbp]
\centering
\begin{subfigure}[t]{0.64\linewidth}
    \vspace{0pt}
    \centering
    \includegraphics[width=\linewidth]{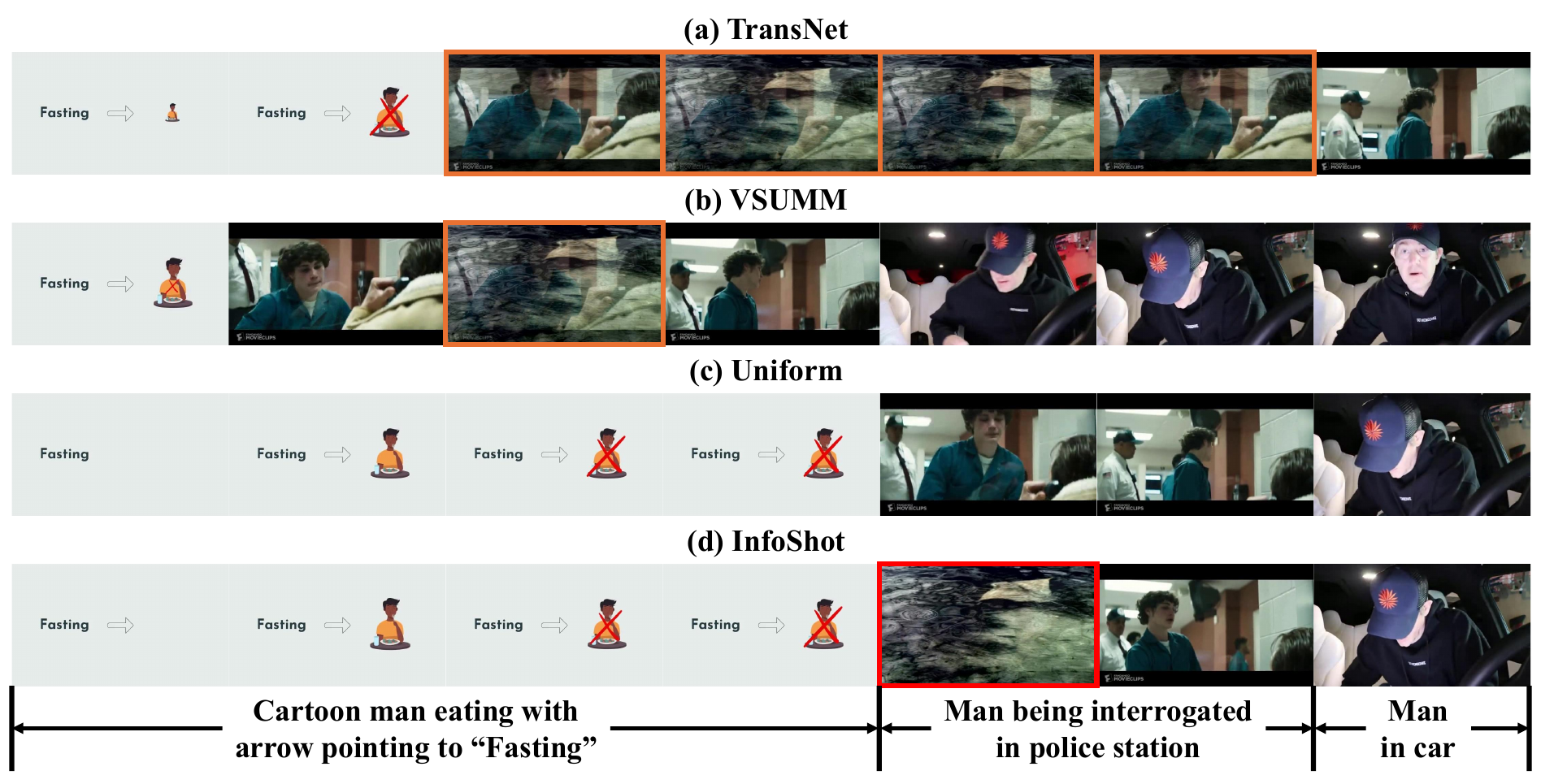}
    \caption{Sampled frames using different methods.}
    \label{fig:case_camou_fig}
\end{subfigure}\hfill
\begin{subfigure}[t]{0.34\linewidth}
    \vspace{0pt}
    \centering
    \includegraphics[width=\linewidth,trim=0 8bp 435bp 0,clip]{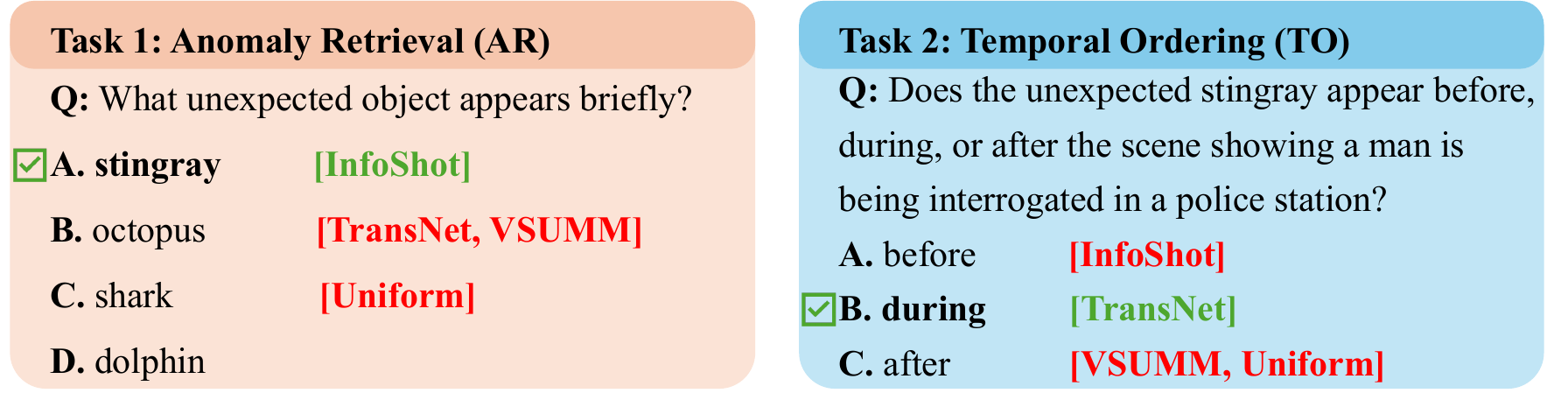}\\[0.3em]
    \includegraphics[width=\linewidth,trim=435bp 8bp 0 0,clip]{figures/case_camou_task.pdf}
    \caption{Video understanding questions and answers selected by the VLM based on different evidence.}
    \label{fig:case_camou_task}
\end{subfigure}
\caption{Case study of camouflage injection pattern.}
\label{fig:case_camou}
\end{figure}

Fig.~\ref{fig:case_camou} shows a camouflage example in which the injected image is color-matched to the local background. For {AR}, InfoShot captures the clearest flash frame (red), enabling the VLM to correctly identify the injected content. TransNet and VSUMM also select frames with partial anomaly evidence (orange), but the visual cues are weaker and insufficient for reliable recognition; Uniform misses the flash frames entirely. For {TO}, TransNet answers correctly in this example when the anomaly is explicitly specified in the prompt. InfoShot also captures the anomaly, but the selected evidence, `stingray', is concentrated earlier than the reference event, `Man in the police station', used in the question, which can lead to an incorrect ordering prediction. This failure case suggests that capturing the anomaly alone is not always sufficient for temporal reasoning; the sampler must also preserve enough surrounding context to establish its relation to the reference event. A promising direction for future work is therefore adaptive budget allocation within a shot, so that more frames can be assigned to temporally ambiguous regions.

\subsection{High-saturation}
\begin{figure}[htbp]
\centering
\begin{subfigure}{\linewidth}
    \centering
    \includegraphics[width=\linewidth]{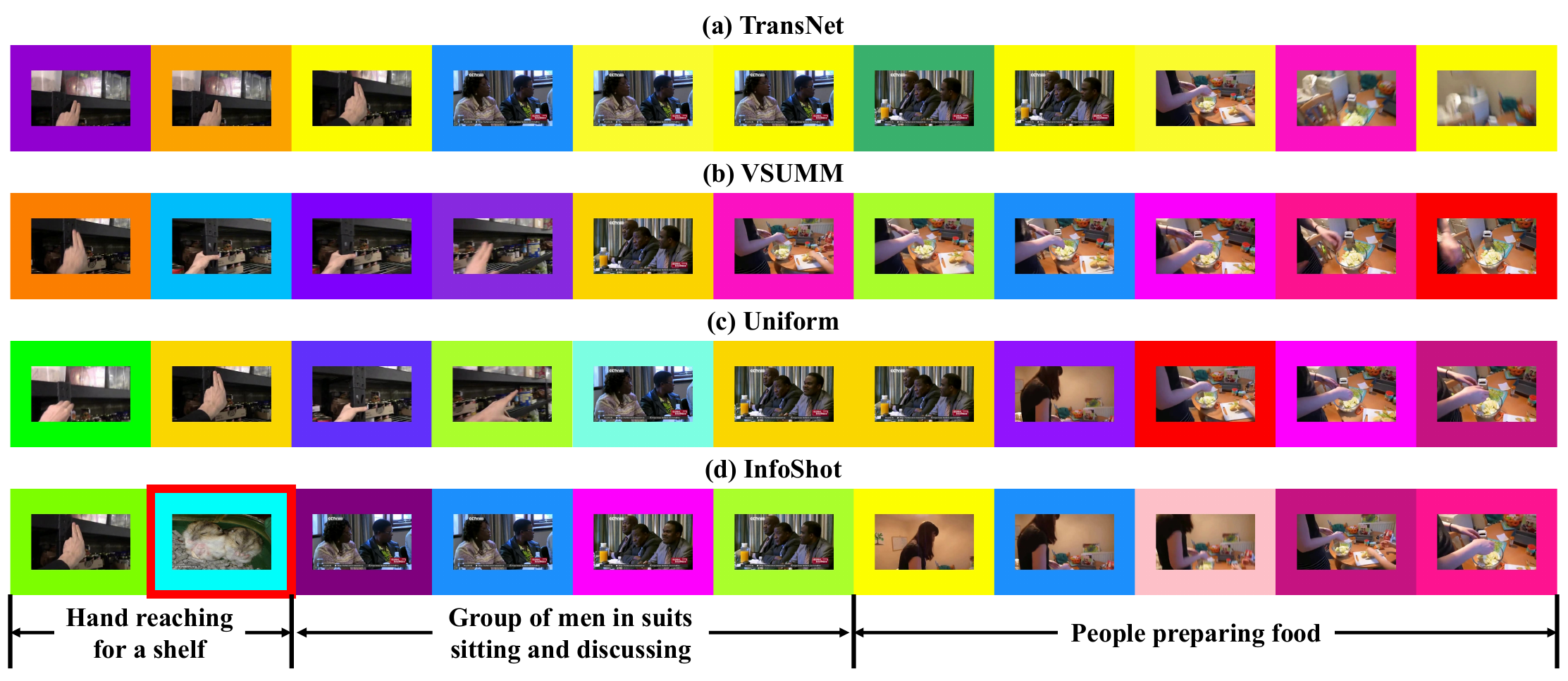}
    \caption{Sampled frames using different methods.}
    \label{fig:case_disco_fig}
\end{subfigure}
\vspace{0.4em}
\begin{subfigure}{\linewidth}
    \centering
    \includegraphics[width=0.8\linewidth]{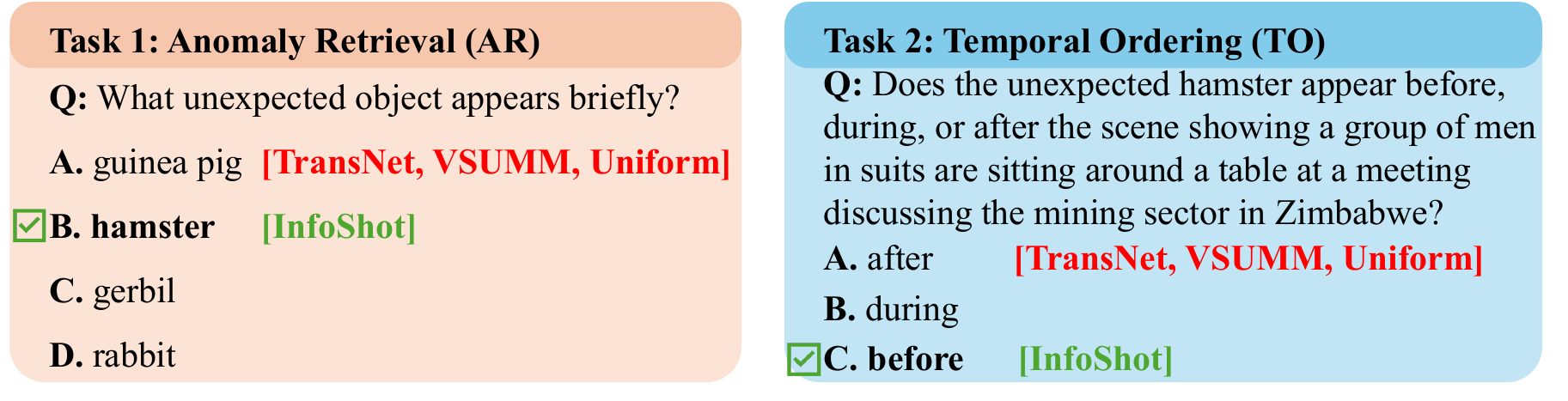}
    \caption{Video understanding questions and answers selected by the VLM based on different evidence.}
    \label{fig:case_disco_task}
\end{subfigure}
\caption{Case study of high-saturation injection pattern.}
\label{fig:case_disco}
\end{figure}

Fig.~\ref{fig:case_disco} shows a high-saturation example with strong background flicker. For {AR}, only InfoShot samples the flash evidence (red), while the other methods are distracted by frequent global color changes and miss the anomaly frames. For {TO}, InfoShot also answers correctly because it preserves both the anomaly frame and a sufficient surrounding context to relate it to the macroscopic event, whereas other methods fail due to missing anomaly evidence.

\Needspace{0.85\textheight}
\subsection{Curtain}
\begin{figure}[!ht]
\centering
\begin{subfigure}{\linewidth}
    \centering
    \includegraphics[width=\linewidth]{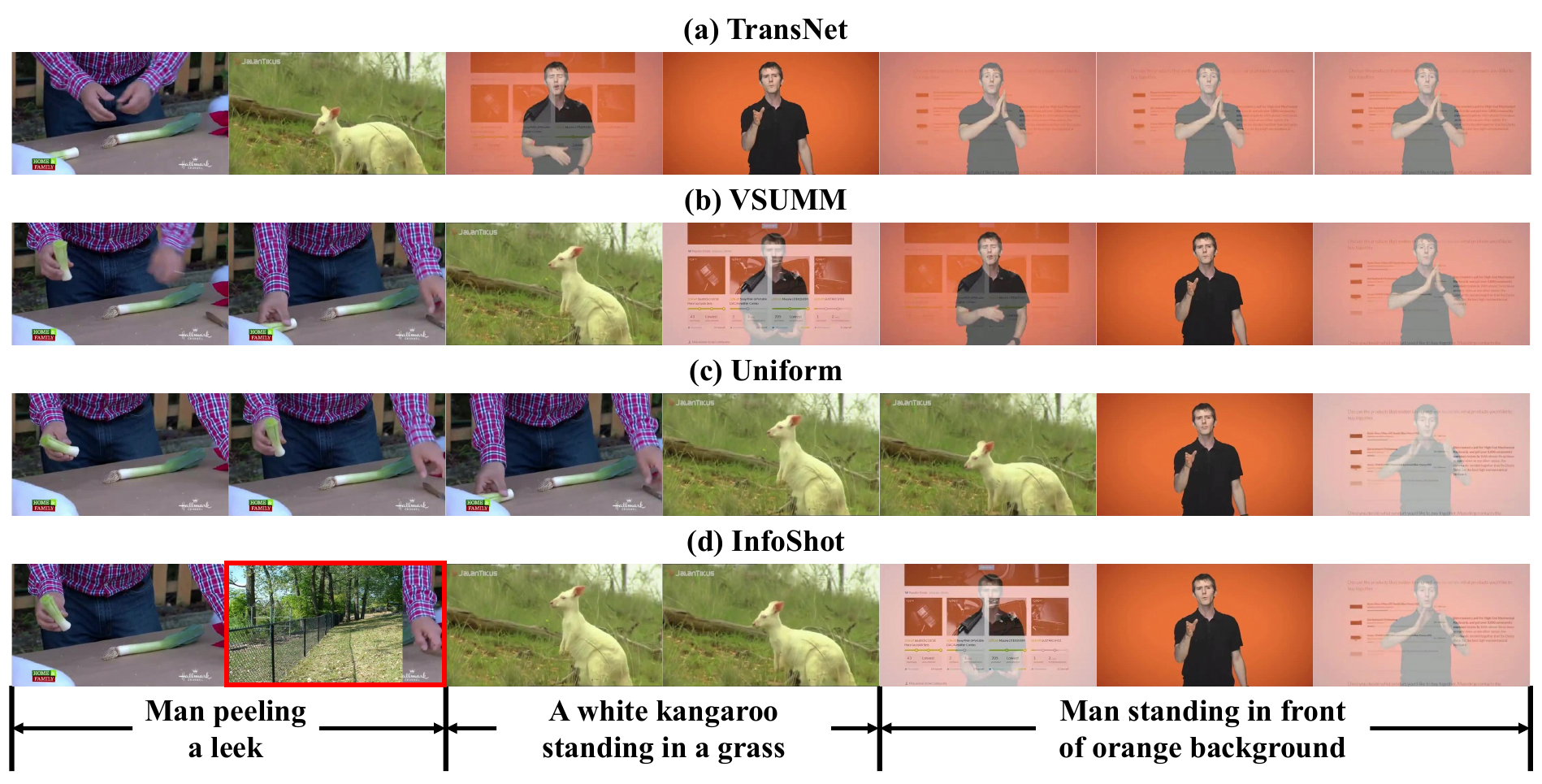}
    \caption{Sampled frames using different methods.}
    \label{fig:case_wipe_fig}
\end{subfigure}
\vspace{0.4em}
\begin{subfigure}{\linewidth}
    \centering
    \includegraphics[width=0.8\linewidth]{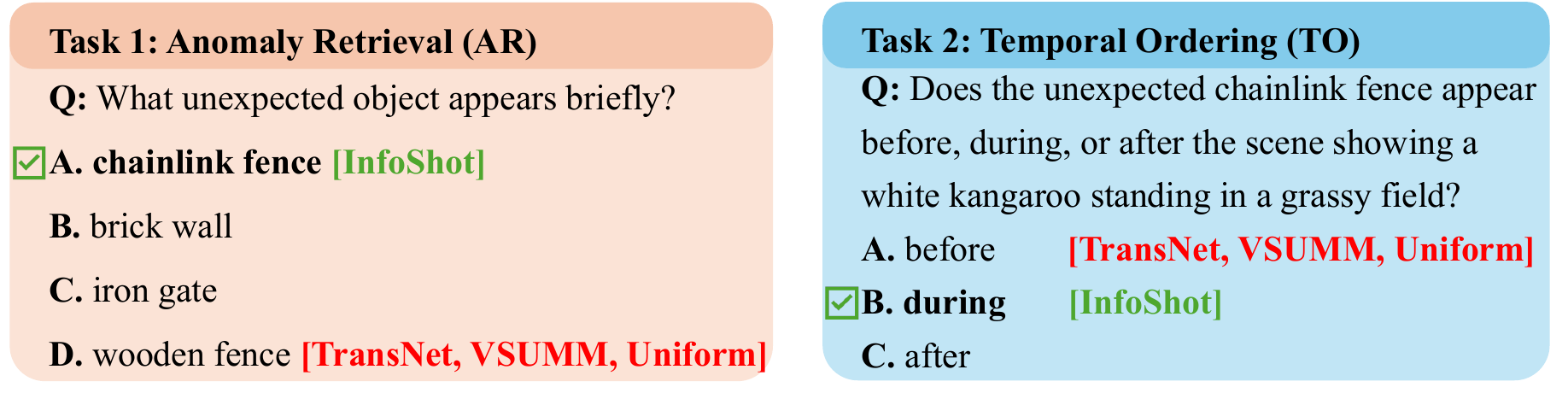}
    \caption{Video understanding questions and answers selected by the VLM based on different evidence.}
    \label{fig:case_wipe_task}
\end{subfigure}
\caption{Case study of curtain injection pattern.}
\label{fig:case_wipe}
\end{figure}

Fig.~\ref{fig:case_wipe} shows a curtain example in which the anomaly is introduced through a short transition. For both \textbf{AR} and \textbf{TO}, InfoShot is the only method that captures frames containing sufficient anomaly information to support correct reasoning, whereas the other samplers miss the key frames.

This pattern also highlights another limitation of the current design. Because the anomaly is revealed progressively during the wipe transition, the most useful frame is not simply any frame near the transition, but the one with the highest information content, i.e., the frame where the injected content is most fully visible. As illustrated in Fig.~\ref{fig:case_wipe}, even when InfoShot succeeds on both tasks, the retained frame still does not fully capture the complete anomaly exposure. This suggests a promising direction for future work: introducing a transition-aware criterion on top of the similarity or affinity structure to prioritize frames with maximal anomaly information, rather than relying only on generic local variation. However, this individual success does not change the aggregate Curtain ranking in main-paper Table~\ref{tab:main_results}.

\Needspace{0.52\textheight}
\subsection{Picture-in-picture (PiP)}
\begin{figure}[!ht]
\centering
\begin{subfigure}{\linewidth}
    \centering
    \includegraphics[width=\linewidth]{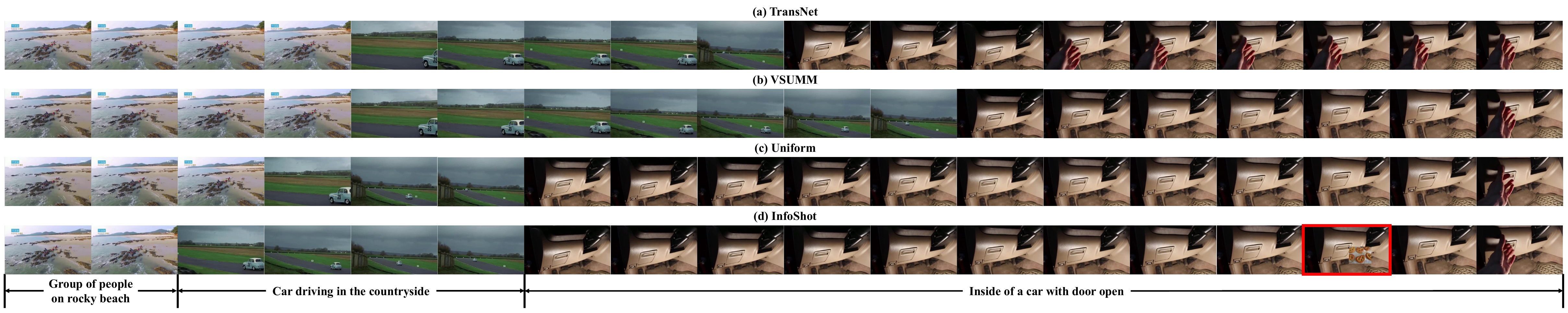}
    \caption{Sampled frames using different methods.}
    \label{fig:case_patch_fig}
\end{subfigure}
\vspace{0.4em}
\begin{subfigure}{\linewidth}
    \centering
    \includegraphics[width=0.8\linewidth]{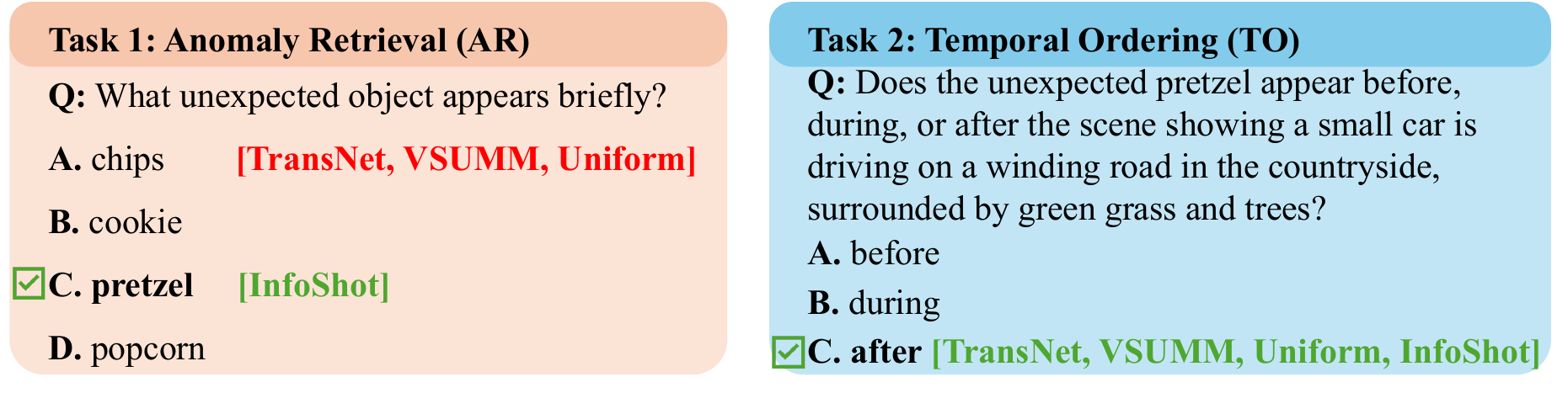}
    \caption{Video understanding questions and answers selected by the VLM based on different evidence.}
    \label{fig:case_patch_task}
\end{subfigure}
\caption{Case study of PiP injection pattern.}
\label{fig:case_patch}
\end{figure}

Fig.~\ref{fig:case_patch} shows a PiP example where the anomaly occupies a small spatial region. For {AR}, InfoShot is the only method that captures the injected patch clearly, while the other methods miss it due to weak global change. For {TO}, although InfoShot is the only sampler that includes the anomaly frame, the VLM can still answer correctly across methods in this example possibly because the question admits a plausible shortcut (the background content is visually similar to the anomaly category), leading to similar predictions despite different evidence.

\end{document}